\documentclass[10pt,twocolumn,letterpaper]{article}

\usepackage{iccv}
\usepackage{times}
\usepackage{epsfig}
\usepackage{graphicx}
\usepackage{amsmath}
\usepackage{amssymb}
\usepackage{makecell}
\usepackage{color}
\usepackage{multirow}
\usepackage{booktabs}
\usepackage{tabularx}
\usepackage[utf8]{inputenc}
\usepackage[toc,page]{appendix}
\usepackage{fancyhdr}

\newcolumntype{C}[1]{>{\centering\let\newline\\\arraybackslash\hspace{0pt}}m{#1}}

\newcommand{\secref}[1]{Section~\ref{#1}}
\renewcommand{\eqref}[1]{Eq.~\ref{#1}}
\newcommand{\tabref}[1]{Table~\ref{#1}}

\newcommand{\refsec}[1]{Section \ref{#1}}

\newcommand\blfootnote[1]{%
  \begingroup
  \renewcommand\thefootnote{}\footnote{#1}%
  \addtocounter{footnote}{-1}%
  \endgroup
}

\makeatletter
\newcommand{\raisemath}[1]{\mathpalette{\raisem@th{#1}}}
\newcommand{\raisem@th}[3]{\raisebox{#1}{$#2#3$}}
\makeatother

\makeatletter
\DeclareRobustCommand\onedot{\futurelet\@let@token\@onedot}
\def\@onedot{\ifx\@let@token.\else.\null\fi\xspace}

\makeatother

\usepackage[pagebackref=true,breaklinks=true,letterpaper=true,colorlinks,bookmarks=false]{hyperref}

\iccvfinalcopy %

\def\iccvPaperID{496} %

\ificcvfinal\fi
\begin{document}

\title{Attacking Optical Flow}

\author{Anurag Ranjan$^\dagger$ \quad Joel Janai$^{\dagger\ddagger}$ \quad Andreas Geiger$^{\dagger\ddagger}$ \quad Michael J. Black$^\dagger$ \\
$^\dagger$Max Planck Institute for Intelligent Systems \\
$^\ddagger$University of T\"ubingen \\
{\tt\small \{aranjan, jjanai, ageiger, black\}@tue.mpg.de}
}

\newcommand{\jj}[1]{#1} %

\maketitle
\thispagestyle{empty}

\begin{abstract}
Deep neural nets achieve state-of-the-art performance on the problem of optical flow estimation. Since optical flow is used in several safety-critical applications like self-driving cars, it is important to gain insights into the robustness of those techniques. Recently, it has been shown that adversarial attacks easily fool deep neural networks to misclassify objects. The robustness of optical flow networks to adversarial attacks, however, has not been studied so far. In this paper, we extend adversarial patch attacks to optical flow networks and show that such attacks can compromise their performance. We show that corrupting a small patch of less than 1\% of the image size can significantly affect optical flow estimates. Our attacks lead to noisy flow estimates that extend significantly beyond the region of the attack, in many cases even completely erasing the motion of objects in the scene. While networks using an encoder-decoder architecture are very sensitive to these attacks, we found that networks using a spatial pyramid architecture are less affected. We analyse the success and failure of attacking both architectures by visualizing their feature maps and comparing them to classical optical flow techniques which are robust to these attacks. We also demonstrate that such attacks are practical by placing a printed pattern into real scenes.

\end{abstract}

\section{Introduction}
Optical flow refers to the apparent 2D motion of each pixel in an image sequence. It is denoted by a vector field $(u,v)$ that corresponds to the displacement of each pixel in the image plane.
The classical formulation \cite{horn1981determining} seeks the optical flow $(u,v)$ between two consecutive images $I(x,y,t)$ and $I(x+u,y+v,t+1)$ in a sequence that minimizes the brightness constancy (i.e.~photometric) error
at each pixel, $\rho(I(x,y,t)- I(x+u,y+v,t+1))$,
for some robust function $\rho$, subject to spatial coherence constraints that regularize the solution \cite{black1996robust}.

\begin{figure}
\begin{center}
\includegraphics[width=\linewidth]{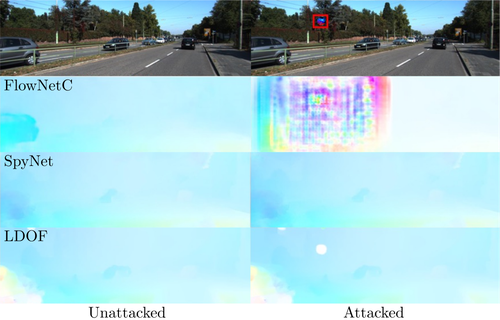}
\end{center}
   \caption{\textbf{Overview}. The first column shows the optical flow results using an encoder-decoder architecture, FlowNetC \cite{dosovitskiy2015flownet}, a spatial pyramid architecture, SpyNet \cite{spynet2017}, and a classical method, LDOF \cite{Brox2011PAMI}. In the second column, a small circular patch is added to both frames at the same location and orientation (highlighted by the red box for illustration purposes). SpyNet and LDOF are barely affected by the patch. In contrast, FlowNetC is strongly affected, even in regions far away from the patch.
   }
\label{fig:teaser}
\end{figure}

The estimation of flow from pairs of images has a long history \cite{horn1981determining} and is used in many applications spanning medicine, special effects, video analysis, action recognition, navigation, and robotics to name a few.
A large number of methods \cite{baker2011database} have approached the problem and steadily improved the results on current benchmarks \cite{Butler:ECCV:2012,Geiger2012CVPR}.
However, classical optical flow methods typically require extensive hand engineering \cite{sun2010secrets} and computationally intensive optimization.
Recent methods \cite{dosovitskiy2015flownet,ilg2017flownet,spynet2017} have therefore approached optical flow estimation using deep neural networks.
These methods typically run in real time and produce results that are competitive with, or even surpass, classical optimization-based approaches. %
Given the performance of recent optical flow networks, they could become an important component in applications such as self-driving cars.
The safety issues surrounding many of these systems implies that the
robustness of the algorithms must be well understood.
To date, there has not been any study to measure the robustness of these networks to adversarial attacks.

Adversarial attacks on neural networks have been shown to work on image classification networks \cite{nguyen2015deep} and, consequently, it is reasonable to ask how such attacks can affect optical flow networks.
Consider the optical flow $(u,v)$ between two frames of an image pair $(I_t, I_{t+1})$ computed using a network $F$ such that $(u,v) = F(I_t, I_{t+1})$.
A successful adversarial attack would cause large changes in estimated optical flow for small, unnoticeable, perturbations in the images.
Specifically, the adversary seeks a perturbed image pair $(\tilde{I_t}, \tilde{I}_{t+1})$ such that \mbox{$\|I_t-\tilde{I_t}\|_0 + \|I_{t+1}-\tilde{I}_{t+1}\|_0<\epsilon$}, where $\epsilon$ is a small constant, and %
the estimated optical flow on the perturbed images is significantly worse, \mbox{$||F(I_t, I_{t+1})- F(\tilde{I_t}, \tilde{I}_{t+1})||>E$}, with a large constant $E$.
We are particularly interested in perturbations under the $\ell_0$ norm,
since the objective is to perturb a very small number of pixels in the original image.

In general, the perturbations on the image can be defined in several ways \cite{goodfellow2014explaining,nguyen2015deep,su2017one,szegedy2013intriguing}.
Here, we focus on applying perturbations by pasting a small patch on the image motivated by
Brown et al.~\cite{brown2017adversarial}, who use such an approach to carry out targeted adversarial attacks on image classification.
The patch attack has real-world significance and we show that adversarial patches can compromise optical flow networks if an engineered patch is printed and placed in real-world scenes.

Optical flow networks can be classified into two types of architectures -- the encoder-decoder \cite{dosovitskiy2015flownet,ilg2017flownet} and spatial pyramid networks \cite{spynet2017,Sun2018CVPR}. We contrast the robustness of adversarial attacks under these two architecture types. Similar to \cite{brown2017adversarial}, we independently and jointly learn patches to attack all networks. The learned patches compromise the encoder-decoder architectures while spatial pyramid networks show more robust behaviour as shown in Figure \ref{fig:teaser}.

In the automotive scenario, cameras for autonomous driving are often behind the windscreen.
Patch attacks can potentially be accomplished by placing the patch on the windshield of the car or placing it in the scene (e.g.~on a traffic sign or other vehicle).
Note that when the patch has zero motion w.r.t.~the camera, classical optical flow algorithms estimate zero optical flow over the patch (LDOF in Fig.~\ref{fig:teaser}).
However, this engineered patch, even if it has no motion, can cause the optical flow predictions from encoder-decoder architectures to be wildly wrong (FlowNetC in Fig.~\ref{fig:teaser}).
In such a scenario, the patch affects the estimated optical flow over large areas in the image that are far away from the patch.
The patch in Figure \ref{fig:teaser} is 0.53\% the size of the image and is barely noticeable, yet it affects the flow in about 50\% of the image region.

For the encoder-decoder networks, the patches not only influence the optical flow estimates in the area of the patch, but their influence extends to remote areas in the image.
In spatial pyramid architectures, in contrast, the patch causes at most small degradations in the area of the patch.
Classical approaches \cite{Brox2011PAMI,Revaud2015CVPR} are even more
robust to adversarial patches.
We propose a \emph{Zero-Flow} test (\secref{sec:analysis}) to analyse the causes of patch attacks, where we visualize the feature maps of the networks while attacking a uniform random noise image without motion. Thereby, we identify three major problems of optical flow architectures. 1) Flow networks are not spatially invariant, leading to spatially varying feature maps even without any motion. 2) Spatial pyramid networks produce large errors at coarse resolutions but are able to recover. 3) Deconvolution layers lead to strong amplification of activations and checkerboard artifacts.

Our contributions are as follows.
We extend adversarial patch attacks to optical
flow neural networks.
We learn adversarial patches and show that these attacks can considerably affect the performance of  optical flow networks based on an encoder-decoder architecture.
We show that spatial pyramid architectures, along with classical optical flow methods, are more robust to patch attacks.
We show that such attacks are easy to implement in practice by simply printing the patch and placing it in the scene.
We also analyse the feature maps of these networks under attack to provide insight into their behaviour under attack.
Code is available at \url{http://flowattack.is.tue.mpg.de/}.

\section{Related Work}

\textbf{Optical Flow.} The classical version of the optical flow problem involves solving for a flow field that minimizes the brightness constancy loss.
A survey of classical optical flow methods is available in \cite{baker2011database,sun2014quantitative}.
These classical methods solve a complex, non-convex, optimization problem and are often slow.
Recent deep learning methods replace the optimization process, and instead directly predict optical flow using convolutional networks \cite{dosovitskiy2015flownet,ilg2017flownet,spynet2017}.
FlowNet \cite{dosovitskiy2015flownet} is the first work to regress optical flow by learning an end-to-end deep neural network.
FlowNet is based on an encoder-decoder architecture with skip connections.
Although FlowNet is much faster than classical methods \cite{sun2010secrets}, it is not as accurate.
More recently, Ilg et al.~\cite{ilg2017flownet} proposed FlowNet2, which achieves state-of-the-art performance on optical flow benchmarks.
FlowNet2 is constructed by stacking multiple FlowNets together and fusing the output with a network specialized on small motions.

In contrast, motivated by the classical coarse-to-fine methods, SpyNet \cite{spynet2017} splits the matching problem into simpler subproblems using coarse-to-fine estimation on an image pyramid.  %
PWC-Net \cite{Sun2018CVPR} extends this idea with a correlation layer to learn optical flow prediction; this gives state-of-the-art results on optical flow benchmarks.

Still other methods approach this problem from the unsupervised learning perspective \cite{meister2018unflow, yu2016back, Janai2018ECCV, ranjan2019competitive}
 by using a neural network to minimize the photometric error %
under certain constraints. These methods use both encoder-decoder architectures \cite{meister2018unflow,yu2016back} and spatial pyramid architectures \cite{Janai2018ECCV}.

While deep networks show impressive results on benchmark datasets, their robustness is not yet well understood.
Robustness, however, is critical for applications such as autonomous driving. Therefore, we investigate the robustness of representative approaches of each category in this paper. In particular, we test FlowNet \cite{dosovitskiy2015flownet}, FlowNet2 \cite{ilg2017flownet}, SpyNet \cite{spynet2017}, PWC-Net \cite{Sun2018CVPR}, and Back2Future \cite{Janai2018ECCV} for their robustness against adversarial attacks.

\textbf{Adversarial Attacks.}
Adversarial attacks seek small perturbations of the input causing large errors in the estimation by a deep neural network.
Attacking neural networks using adversarial examples is a popular way to examine the reliability of networks for image classification \cite{goodfellow2014explaining,nguyen2015deep,szegedy2013intriguing}.
The key to all such attacks is that the change to the image should be minor, yet have a large influence on the output.
Adversarial examples typically involve small perturbations to the image that are not noticeable to the human eye.
The adversaries are shown to work even when a single pixel is perturbed in the image \cite{su2017one}.

Although these attacks reveal limitations of deep networks, they can not be easily replicated in real-world settings.
For instance, it is rather difficult to change a scene such that one pixel captured by a camera is perturbed in a specific way to fool the network.
However, recent work \cite{kurakin2016adversarial} demonstrates that adversarial examples can also work when printed out and shown to the network under different illumination conditions.
Athalye et al.~\cite{athalye2017synthesizing} show that adversarial examples can be 3D printed and are misclassified by networks at different scales and orientations.
Sharif et al.~\cite{sharif2016accessorize} construct adversarial glasses to fool facial recognition systems.
Etimov et al.~\cite{evtimov2017robust} show that stop signs can be misclassified by placing various stickers on top of them.

Recently proposed patch attacks \cite{brown2017adversarial} place a small engineered patch in the scene to fool the network into making wrong decisions. Patch attacks are interesting because they can be easily replicated in the real world, and work at several scales and orientations of the patch. This makes deep networks vulnerable in real world applications. Therefore, we focus our investigation of the robustness of deep flow networks on these kinds of attacks.

To our knowledge, there has been no work on attacking optical flow networks.
Consequently, we explore how such networks can be attacked using adversarial patches and analyse the potential causes of such vulnerabilities.

\section{Approach}
Adversarial attacks are carried out by optimizing for a perturbation that forces a network to output the wrong labels compared with ground truth labels.
For example, if $(u,v)$ represents the ground truth labels for inputs $(I_t, I_{t+1})$, a perturbed input $(\tilde{I}_t, \tilde{I}_{t+1})$ would produce incorrect labels $(\tilde{u}, \tilde{v})$.
However, there are no optical flow datasets that have dense optical flow ground truth labels for natural images. Most of the optical flow datasets are synthetic \cite{Butler:ECCV:2012, dosovitskiy2015flownet, ranjan2018learning}.
SlowFlow \cite{janai2017slow} provides real world data but is limited in size.
The KITTI dataset \cite{Geiger2012CVPR} has sparse ground truth labels and the annotations are limited to 200 training examples.
To address the problem of limited ground truth labels, we use the predictions of the optical flow network as pseudo ground truth. We optimize for the perturbation that maximizes the angle between the predicted flow vectors obtained using the original images and the perturbed images respectively. Using the predictions instead of ground truth has the advantage that a patch can be optimized using any unlabelled video. This makes it easier to attack optical flow systems even in the absense of the ground truth.

Consider an optical flow network $F$ that computes the optical flow between two-frames of an image sequence $(I_t, I_{t+1})$ of resolution $H \times W$. Consider a small patch $p$ of resolution $h \times w$ that is pasted onto the image to perturb it.
Let $\delta \in \mathcal{T}$ be a transformation that can be applied to the patch. These transformations in $\mathcal{T}$ can be a combination of rotations and scaling. We define the perturbation $A(I,p,\delta,l)$ on the image $I$, that applies the transformations $\delta$ to the patch $p$ and pastes it at a location $l \in \mathcal{L}$ in the image. We apply the same perturbation to both frames in the sequence $\tilde{I}_t = A(I_t, p, \delta,l)$ and $\tilde{I}_{t+1} = A(I_{t+1}, p, \delta,l)$. This means that the optical flow between the perturbed frames is zero in the region containing the patch.
In the real world, this would correspond to a patch being stationary with respect to the camera. In our experiments, we show that patches obtained with this assumption generalize well to the realistic scenario of moving patches where $\tilde{I}_t = A(I_t, p, \delta_t,l_t)$ and $\tilde{I}_{t+1} = A(I_{t+1}, p, \delta_{t+1},l_{t+1})$ with $\delta_t \neq \delta_{t+1}$ and $l_{t} \neq l_{t+1}$. %

Our goal is to learn a patch $p$ that acts as an adversary to an optical flow network, $F$, and is invariant to location $l$ or transformations $\delta$ of the patch. %
The resulting patch can be optimized using
\begin{equation}
\label{eq:adv}
\hat{p} = \operatorname*{argmin}_p \mathbb{E}_{(I_t,I_{t+1}) \sim \mathcal{I}, l \sim \mathcal{L},\delta \sim \mathcal{T}} \frac{(u,v) \cdot (\tilde{u}, \tilde{v})}{\|(u,v)\| \cdot \|(\tilde{u}, \tilde{v})\|}
\end{equation}
with
\begin{align}
(u,v) &= F(I_t, I_{t+1})  \\
(\tilde{u}, \tilde{v}) &= F(A(I_t,p,\delta,l), A(I_{t+1},p,\delta,l))
\end{align}
where $l$ is sampled over all the locations $\mathcal{L}$ in the image, and $\mathcal{I}$ is a set of 2 frame sequences from a video.

Equation (\ref{eq:adv}) computes the cosine of the angle between the optical flow estimated by the network for normal and perturbed images.
Therefore, minimizing the loss is equivalent to finding those adversarial examples that reverse the direction of optical flow estimated by the network.

\section{Experiments}
\begin{figure}[t]
\begin{center}
{\def\arraystretch{0}\tabcolsep=0pt
\begin{tabular}{@{}p{0.15\linewidth}@{}C{0.15\linewidth}@{}C{0.2\linewidth}@{}C{0.25\linewidth}@{}C{0.3\linewidth}@{}}
	\rotatebox[origin=c]{40}{FlowNetC} & 
	\includegraphics[width=0.5\linewidth]{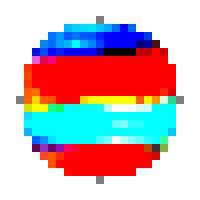} &
	\includegraphics[width=0.6\linewidth]{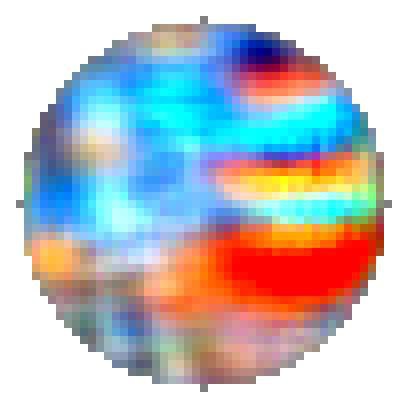} & 
	\includegraphics[width=0.7\linewidth]{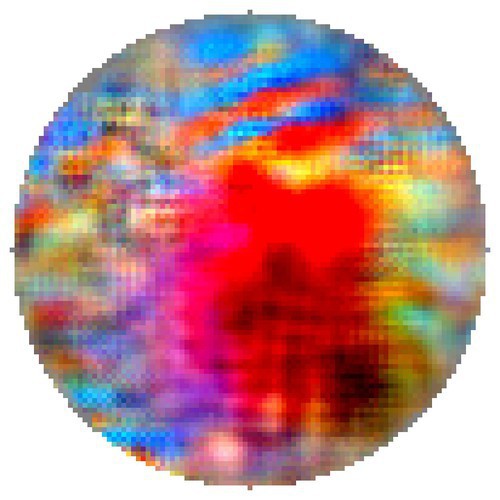} & 
	\includegraphics[width=0.8\linewidth]{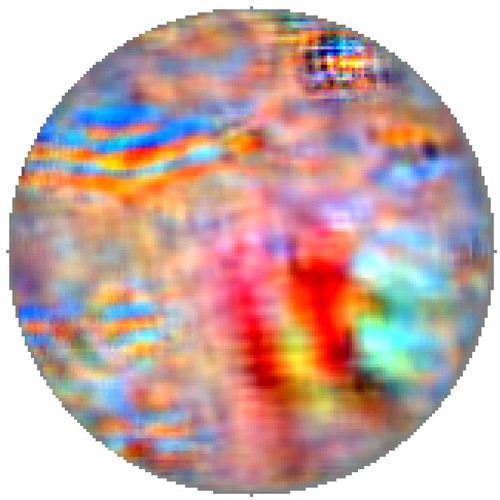} \\ 
	\rotatebox[origin=c]{40}{FlowNet2} & 
	\includegraphics[width=0.5\linewidth]{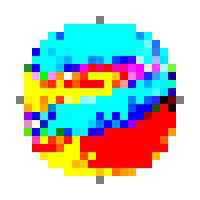} &
	\includegraphics[width=0.6\linewidth]{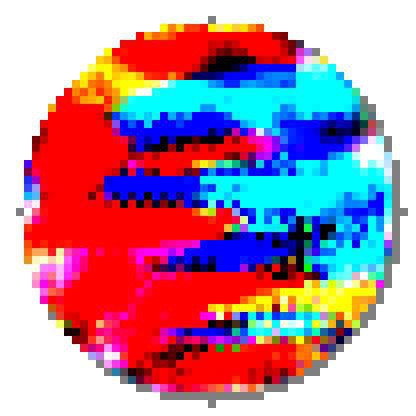} & 
	\includegraphics[width=0.7\linewidth]{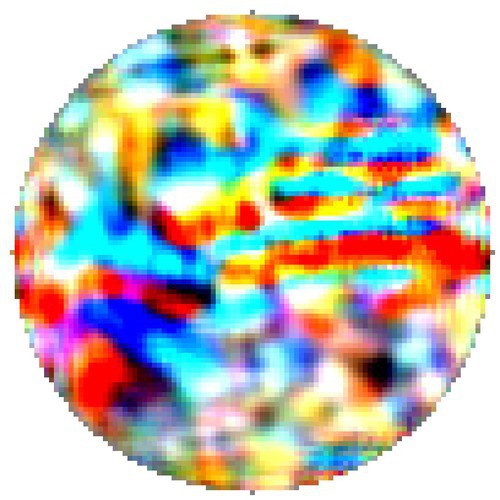} & 
	\includegraphics[width=0.8\linewidth]{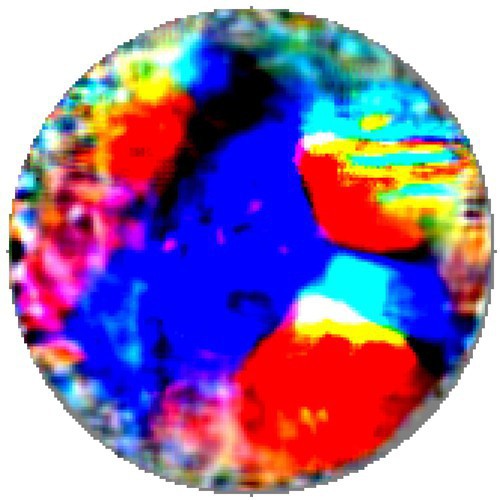} \\
	\rotatebox[origin=c]{40}{SpyNet} & 
	\includegraphics[width=0.5\linewidth]{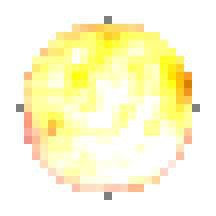} &
	\includegraphics[width=0.6\linewidth]{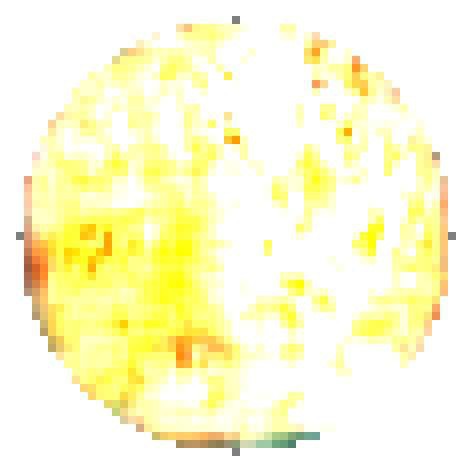} & 
	\includegraphics[width=0.7\linewidth]{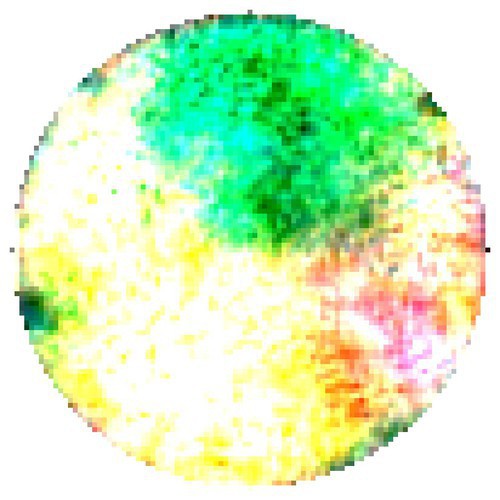} & 
	\includegraphics[width=0.8\linewidth]{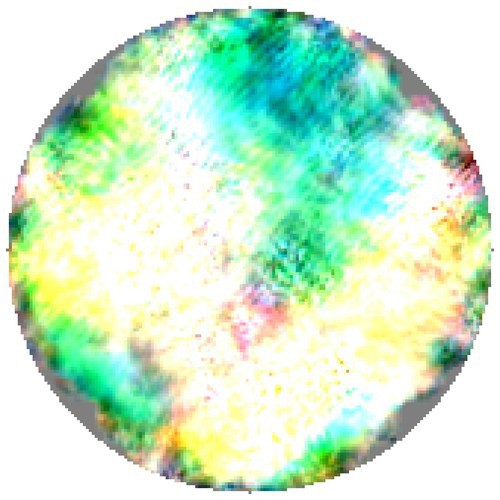} \\ 
	\rotatebox[origin=c]{40}{PWC-Net} & 
	\includegraphics[width=0.5\linewidth]{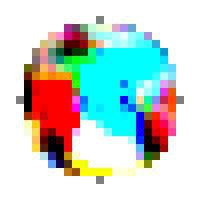} &
	\includegraphics[width=0.6\linewidth]{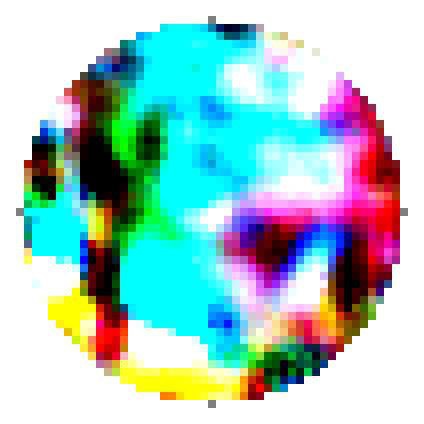} & 
	\includegraphics[width=0.7\linewidth]{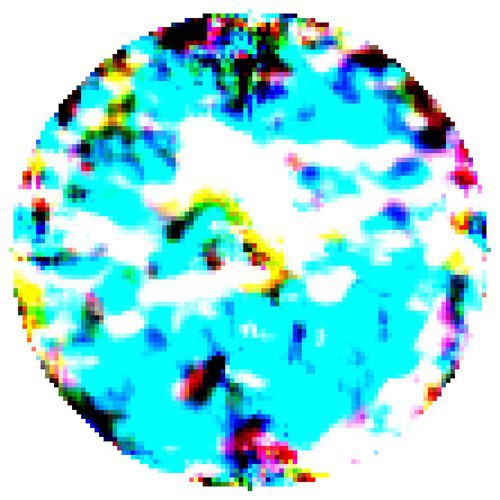} & 
	\includegraphics[width=0.8\linewidth]{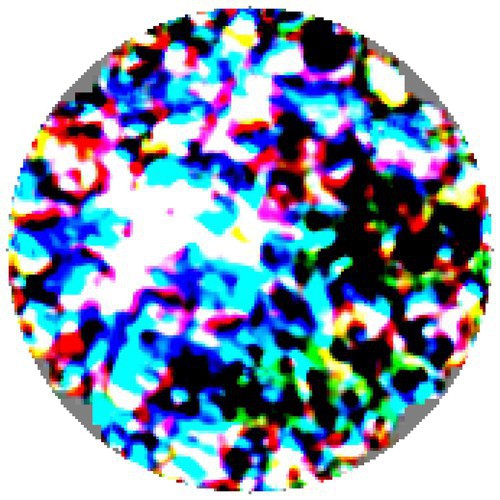} \\
	\rotatebox[origin=c]{40}{Back2Future}  & 
	\includegraphics[width=0.5\linewidth]{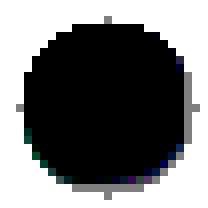} &
	\includegraphics[width=0.6\linewidth]{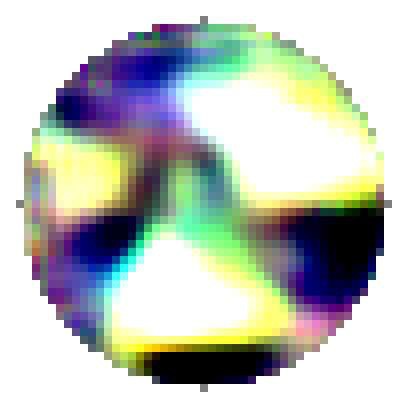} & 
	\includegraphics[width=0.7\linewidth]{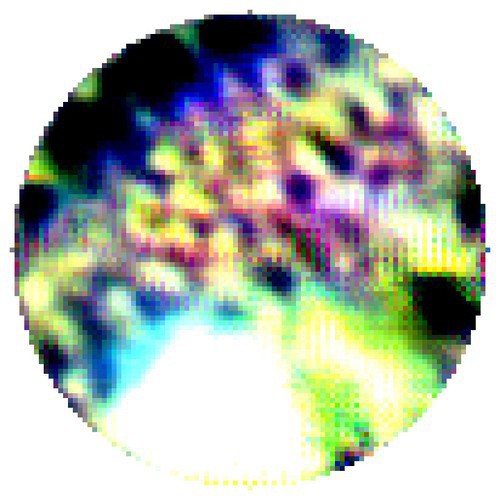} & 
	\includegraphics[width=0.8\linewidth]{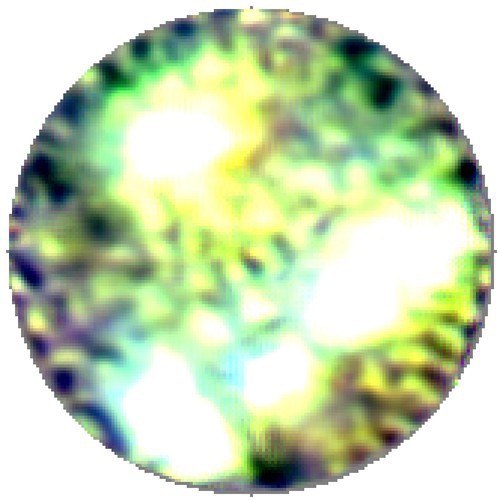} \\
&&&&\\[0.5em]
& 25x25 & 51x51 & 102x102 & 153x153
\end{tabular}
}
\end{center}
\caption{\textbf{Adversarial Patches}. Obtained for different optical flow
  networks. The size is enlarged for visualization purposes.}
\label{fig:all_patches}
\end{figure}

\begin{table}[t]
	\begin{center}
		\caption{\textbf{Optical Flow Methods.} Contrasting different optical flow methods with respect to the Type, Supervision (Super.), Network Architecture (Net.) - Encoder-Decoder (ED) vs.~Spatial Pyramid (SP) - and Number of Parameters (\#Params.). Number of parameters is denoted in Millions (M).}
		\begin{tabular}{lcccc}
			& \textbf{Type} & \textbf{Super.} & \textbf{Net.} & \# \textbf{Params}\\ \hline
			FlowNetC \cite{dosovitskiy2015flownet} & CNN & Yes & ED  & 39 M \\
			FlowNet2 \cite{ilg2017flownet} & CNN & Yes & ED & 162 M \\
			SpyNet \cite{spynet2017} & CNN & Yes & SP & 1.2 M \\
			PWC-Net \cite{Sun2018CVPR} & CNN & Yes & SP & 8.75 M\\
			Back2Future \cite{Janai2018ECCV} & CNN & No  & SP & 12.2 M \\
			LDOF \cite{Brox2011PAMI} & Classic & -  &  - & -\\
			Epic Flow \cite{Revaud2015CVPR} & Classic & - & -  & - \\
		\end{tabular}
		\label{tab:flow_methods}
		\vspace{-0.2cm}
	\end{center}
\end{table}

We evaluate the robustness of five optical networks to adversarial patch attacks -- FlowNetC \cite{dosovitskiy2015flownet}, FlowNet2 \cite{ilg2017flownet}, SpyNet \cite{spynet2017}, PWC-Net \cite{Sun2018CVPR} and Back2Future \cite{Janai2018ECCV}.
We also evaluate the robustness of two very different classical methods -- LDOF \cite{Brox2011PAMI} and EpicFlow \cite{Revaud2015CVPR} which take image derivatives (LDOF) and sparse matches (EpicFlow) to compute optical flow.
These methods cover deep networks vs.~classical methods, supervised vs.~unsupervised learning methods and encoder-decoder vs.~spatial pyramid architectures, thus providing an overall picture of optical flow methods under adversarial patch attacks.
We contrast these optical flow methods in Table \ref{tab:flow_methods}. %
First, we evaluate the effect of \textit{White-box} attacks, where we learn a patch of a specific size for each network by optimizing over that particular network. We then show \textit{Black-box} attacks by creating a universal patch using two different networks and testing all the optical flow methods on this patch in a real world scenario.

\begin{table*}[t]
\begin{center}
\caption{\textbf{White-box Attacks.} Effect of patch size (in pixel and percent of the image size) on different optical flow networks. We show average End Point Error (EPE) on KITTI 2015 for each network with and without an attack. For attacks, we also show relative degradation \% in the EPE.}
\begin{tabular}{l|c|cc|cc|cc|cc}
      &  Unattacked & \multicolumn{2}{c|}{25x25 (0.1\%)} & \multicolumn{2}{c|}{51x51 (0.5\%)} & \multicolumn{2}{c|}{102x102 (2.1\%)} & \multicolumn{2}{c}{153x153 (4.8\%)} \\
    Network & EPE & EPE & Rel & EPE & Rel & EPE & Rel & EPE & Rel \\
\hline
    FlowNetC \cite{dosovitskiy2015flownet} &       14.56 &  29.07 &  +100\% &  40.27 &  +177\% &    82.41 &  +466\% &    95.32 &  +555\% \\
    FlowNet2 \cite{ilg2017flownet} &       11.90 &  17.04 &   +43\% &  24.42 &  +105\% &    38.57 &  +224\% &    59.58 &  +400\% \\
      SpyNet \cite{spynet2017} &       20.26 &  20.59 &    +2\% &  21.00 &    +4\% &    21.22 &    +5\% &    21.00 &    +4\% \\
      PWC-Net \cite{Sun2018CVPR} &       11.03 &  11.37 &    +3\% &  11.50 &    +4\% &    11.86 &    +7\% &    12.52 &   +13\% \\
 Back2Future \cite{Janai2018ECCV} &       17.49 &  18.04 &    +3\% &  18.24 &    +4\% &    18.73 &    +7\% &    18.43 &    +5\% \\
\end{tabular}

\label{tab:sizevepe}
\end{center}
\end{table*}

\subsection{White-box Attacks}
\label{subsec:white_box}
We perform White-box attacks on each of the networks independently. We learn a circular patch that is invariant to its location in the image, scale and orientation. The scale augmentations of the patch are kept within $\pm 5$\% of the original size, and rotations are varied within $\pm 10^{\circ}$. For each network, we learn patches of four different sizes.
The patch size is kept under $5$\% the size of the images being attacked. We optimize for each patch $p$ using Eq.~(\ref{eq:adv}) on unlabeled frames from the raw KITTI \cite{Geiger2012CVPR}.
We choose the KITTI dataset since it reflects a safety critical application and provides an annotated training set for evaluation.
We use the optical flow predictions from the networks as pseudo ground truth labels to optimize for the patch that produces the highest angular error. This allows us to leverage approximately 32000 frames instead of 200 annotated frames from the training dataset.
For each of the networks, we use the pre-trained model that gives the best performance \textit{without} fine tuning on KITTI dataset.
We use Pytorch \cite{pytorch} as our optimization framework and optimize for the patch using stochastic gradient descent.
Figure \ref{fig:all_patches} shows the patches we obtain by optimizing over different optical flow networks and different patch sizes.

\paragraph{Evaluation.}
To quantify the robustness of the networks to our adversarial patch attacks, we measure their performance on the training set of the KITTI 2015 optical flow benchmark using the original and perturbed images, respectively.
We measure the vulnerability of the optical flow networks using average end point error (EPE) and relative degradation of EPE in the presence of the adversarial patch (Table~\ref{tab:sizevepe}). The errors are computed by placing the learned patch $p$ at a random location for each image in the training set.
All images are resized to $384 \times 1280$ and input to the networks. We evaluate optical flow at the full image resolution.

As shown in Table \ref{tab:sizevepe}, the performance of encoder-decoder architectures -- FlowNetC and FlowNet2 degrades significantly, even when using very small patches ($0.1$\% of the image size).
The increase in error is about $400-500$\% under attack.
For the spatial pyramid based methods -- SpyNet, PWC-Net and Back2Future, the degradation is small.
Back2Future, which is an unsupervised method and shares a similar architecture with PWC-Net, suffers less degradation than PWC-Net on the largest patch size.

\begin{figure*}[t]
\renewcommand{\arraystretch}{0}%
\begin{center}

\begin{tabular}{@{}p{0.100\linewidth}@{}C{0.225\linewidth}@{}C{0.225\linewidth}@{}C{0.225\linewidth}@{}C{0.225\linewidth}@{}}
\rotatebox[origin=c]{30}{FlowNetC} & 
 \includegraphics[width=\linewidth]{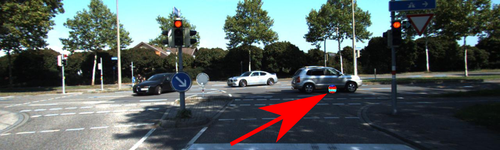} & 
 \multicolumn{3}{@{}m{\linewidth}@{}}{ \includegraphics[width=0.675\linewidth]{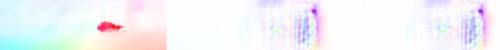}}\\ 
\rotatebox[origin=c]{30}{FlowNet2} & 
 \includegraphics[width=\linewidth]{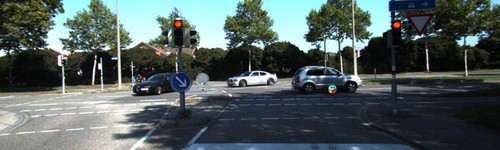} & 
 \multicolumn{3}{@{}m{\linewidth}@{}}{ \includegraphics[width=0.675\linewidth]{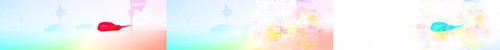}}\\ 
\rotatebox[origin=c]{30}{SpyNet} & 
 \includegraphics[width=\linewidth]{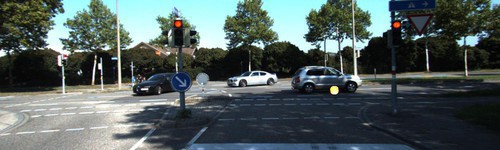} & 
 \multicolumn{3}{@{}m{\linewidth}@{}}{ \includegraphics[width=0.675\linewidth]{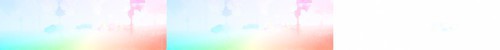}}\\ 
\rotatebox[origin=c]{30}{PWC-Net} & 
 \includegraphics[width=\linewidth]{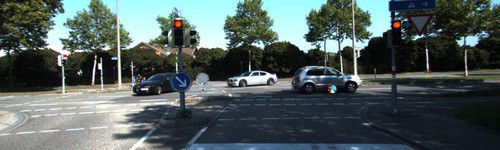} & 
 \multicolumn{3}{@{}m{\linewidth}@{}}{ \includegraphics[width=0.675\linewidth]{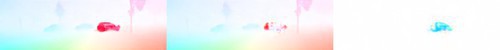}}\\ 
\rotatebox[origin=c]{30}{Back2Future} & 
 \includegraphics[width=\linewidth]{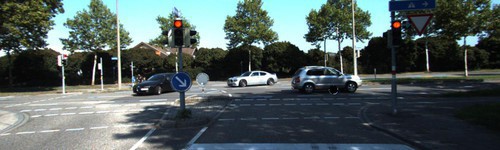} & 
 \multicolumn{3}{@{}m{\linewidth}@{}}{ \includegraphics[width=0.675\linewidth]{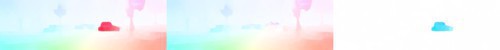}}\\ 
 &&&&\\[0.5em]
& Attacked Reference & Unattacked Flow & Attacked Flow & Difference
\end{tabular}
\end{center}
\caption{\textbf{White-box Attacks.} 25x25 patch attacks on all networks.}
\label{fig:25_patch_attack}
\end{figure*}

\begin{figure*}[t]
\renewcommand{\arraystretch}{0}%
\begin{center}

\begin{tabular}{@{}p{0.100\linewidth}@{}C{0.225\linewidth}@{}C{0.225\linewidth}@{}C{0.225\linewidth}@{}C{0.225\linewidth}@{}}
\rotatebox[origin=c]{30}{FlowNetC} & 
 \includegraphics[width=\linewidth]{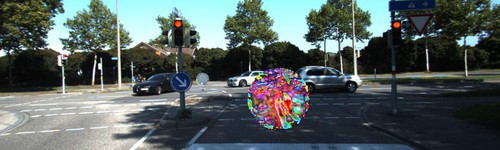} & 
 \multicolumn{3}{@{}m{\linewidth}@{}}{ \includegraphics[width=0.675\linewidth]{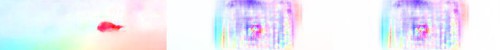}}\\ 
\rotatebox[origin=c]{30}{FlowNet2} & 
 \includegraphics[width=\linewidth]{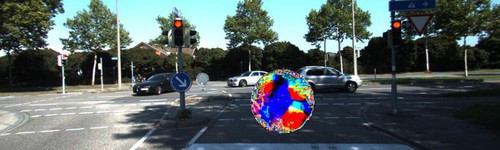} & 
 \multicolumn{3}{@{}m{\linewidth}@{}}{ \includegraphics[width=0.675\linewidth]{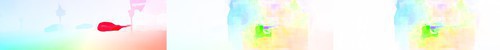}}\\ 
\rotatebox[origin=c]{30}{SpyNet} & 
 \includegraphics[width=\linewidth]{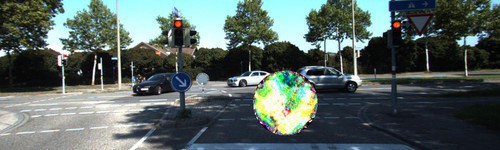} & 
 \multicolumn{3}{@{}m{\linewidth}@{}}{ \includegraphics[width=0.675\linewidth]{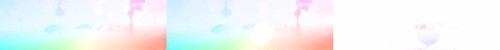}}\\ 
\rotatebox[origin=c]{30}{PWC-Net} & 
 \includegraphics[width=\linewidth]{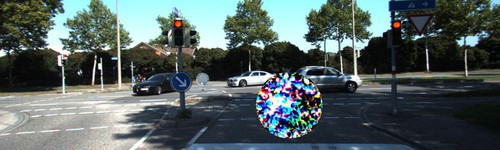} & 
 \multicolumn{3}{@{}m{\linewidth}@{}}{ \includegraphics[width=0.675\linewidth]{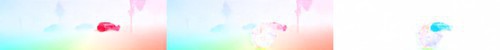}}\\ 
\rotatebox[origin=c]{30}{Back2Future} & 
 \includegraphics[width=\linewidth]{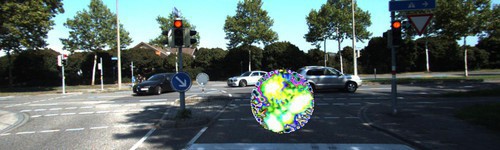} & 
 \multicolumn{3}{@{}m{\linewidth}@{}}{ \includegraphics[width=0.675\linewidth]{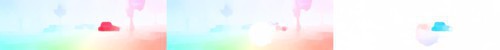}}\\ 
&&&&\\[0.5em]
& Attacked Reference & Unattacked Flow & Attacked Flow & Difference
\end{tabular}
\end{center}
\caption{\textbf{White-box Attacks.} 153x153 patch attacks on all networks.}
\label{fig:153_patch_attack}
\end{figure*}

In Figure \ref{fig:25_patch_attack}, we illustrate that encoder-decoder architectures are extremely vulnerable to patch attacks irrespective of scene content, location and orientation of the patch.
The patch size in Figure \ref{fig:25_patch_attack} is $25 \times 25$ pixels, about $0.12$\% of the size of the image being attacked.
The figure  shows that the attacks extend significantly beyond the region of the patch,  causing a significant proportion of the optical flow vectors to be degraded.
We also observe that SpyNet is least affected by the attack. For PWC-Net and Back2Future, the degradation is contained within the region of the patch. The strength of the attack increases with the size of the patch.
Figure \ref{fig:153_patch_attack} shows attacks with a larger patch size of $153 \times 153$, which is about $4.76$\% of the image size.
Other examples and patch sizes are shown in the supplementary material.

\begin{table}[t]
\begin{center}
\caption{\textbf{Black-box Attacks.} Attacks on different optical flow methods using a universal patch. Methods below the line were not used for training the patch. %
}
\begin{tabular}{l|c|ccc}
	& Unattacked & \multicolumn{2}{c}{Attacked}    \\
	&  EPE   &EPE  &  Rel\\
	\hline
         FlowNet2 \cite{ilg2017flownet} &       11.90 &        36.13 &     +203 \% \\
              PWC-Net \cite{Sun2018CVPR} &       11.03 &        11.01 &      +0 \% \\
              \hline
  FlowNetC \cite{dosovitskiy2015flownet} &     14.56 &      86.12 &     +492 \% \\
  				SpyNet \cite{spynet2017} &       20.26 &        20.39 &       +1 \% \\
       Back2Future \cite{Janai2018ECCV} &       17.49 &        17.44 &      +0 \% \\
        Epic Flow \cite{Revaud2015CVPR} &        4.52 &         4.66 &       +3 \% \\
               LDOF \cite{Brox2011PAMI} &        9.20 &         9.17 &      +0 \% \\
\end{tabular}

\label{tab:epeuni}
\vspace{-0.7cm}
\end{center}
\end{table}

\subsection{Black-box Attacks}
\label{sec:blackbox_attacks}
In a real world scenario, the network used by a system like an autonomous car will most likely not be accessible to optimize the adversarial patch.
Therefore, we also consider a Black-box attack, which learns a ``universal'' patch to attack a group of networks.
Since there are two architecture types (Encoder-Decoder and Spatial Pyramid), we consider one network from each type to learn a ``universal'' patch to attack all networks.
We pick the networks -- FlowNet2 \cite{ilg2017flownet}  and PWC-Net \cite{Sun2018CVPR} %
and jointly optimize for a patch that attacks both networks using Eq.~\ref{eq:adv}. In this way, we obtain a patch that is capable of attacking both encoder-decoder and spatial pyramid architectures. The resulting patch can be seen in Figure \ref{fig:real_flownetc} and the Sup.~Mat.
While we focus on two networks for a proof of concept, the  ``universal'' patch can easily be jointly optimized using more networks to obtain even more effective attacks.

The training of the universal patch is identical to the White-box case. However, we decided to make the evaluation of the Black-box attacks more realistic by using the camera motion and disparity ground truth provided by the KITTI Raw dataset to compute realistic motion of patches assuming they are part of the static scene. We project the random location of the patch from the first frame into the 3D scene and re-project it into the second frame considering the camera motion and disparity. We randomly pick a disparity between the maximal disparity of the whole scene and minimal disparity in region of the patch to obtain a location in the free space between the camera and other objects in the scene. Given the re-projection of points on the patch, we then estimate a homography to transform the patch and warp it using bilinear interpolation. For sequences without ground truth camera poses we use zero motion like in the White-box attacks. In this way, we simulate the situation where a patch is attached to some static object in the scene. In the supplementary material, we also show the effect of a universal patch that has zero motion w.r.t the camera as in the White-box attacks.

\paragraph{Evaluation.} %
Table \ref{tab:epeuni} shows the performance of all networks and classical optical flow methods from Table \ref{tab:flow_methods} in the presence of the Black-box adversary.
Consistent with White-box attacks, we observe that encoder-decoder architectures suffer
significantly in the presence of the adversarial patch.
Both, spatial pyramid architectures and classical methods, are more robust to patch attacks.

\begin{figure*}[t]
	\renewcommand{\arraystretch}{0}%
	\begin{center}
	
\begin{tabular}{@{}p{0.100\linewidth}@{}C{0.300\linewidth}@{}C{0.300\linewidth}@{}C{0.300\linewidth}@{}}
\rotatebox[origin=c]{30}{Inputs / GT} & \includegraphics[width=\linewidth]{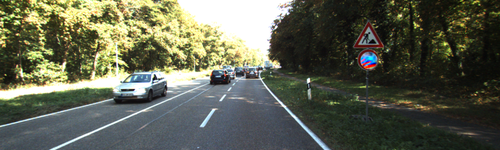} &\includegraphics[width=\linewidth]{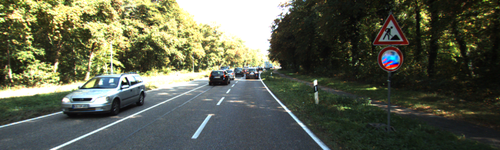} &\includegraphics[width=\linewidth]{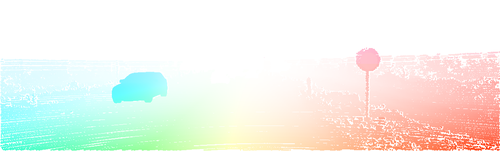} \\ 
 & & & \\[0.5em] \\ 
\rotatebox[origin=c]{30}{FlowNetC} & \multicolumn{3}{@{}m{\linewidth}@{}}{ \includegraphics[width=0.900\linewidth]{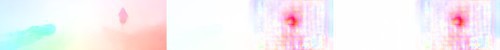}}\\ 
\rotatebox[origin=c]{30}{FlowNet2} & \multicolumn{3}{@{}m{\linewidth}@{}}{ \includegraphics[width=0.900\linewidth]{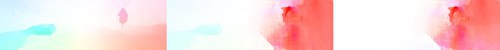}}\\ 
\rotatebox[origin=c]{30}{SpyNet} & \multicolumn{3}{@{}m{\linewidth}@{}}{ \includegraphics[width=0.900\linewidth]{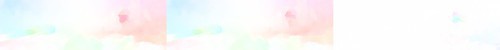}}\\ 
\rotatebox[origin=c]{30}{PWC-Net} & \multicolumn{3}{@{}m{\linewidth}@{}}{ \includegraphics[width=0.900\linewidth]{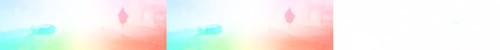}}\\ 
\rotatebox[origin=c]{30}{Back2Future} & \multicolumn{3}{@{}m{\linewidth}@{}}{ \includegraphics[width=0.900\linewidth]{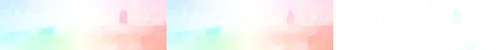}}\\ 
\rotatebox[origin=c]{30}{Epic Flow} & \includegraphics[width=\linewidth]{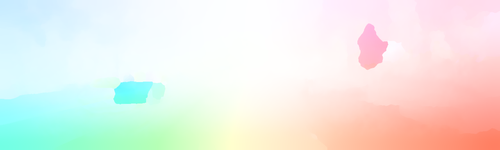} &\includegraphics[width=\linewidth]{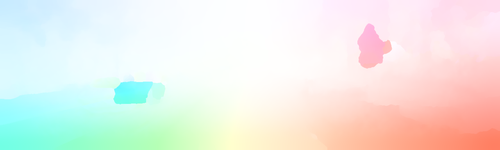} &\includegraphics[width=\linewidth]{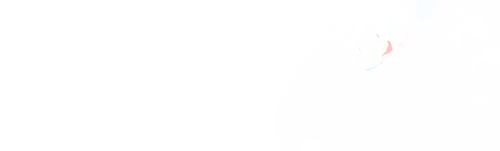} \\ 
\rotatebox[origin=c]{30}{LDOF} & \includegraphics[width=\linewidth]{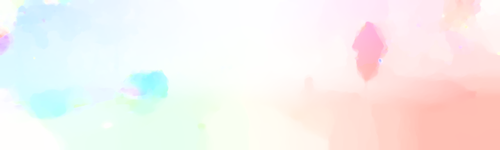} &\includegraphics[width=\linewidth]{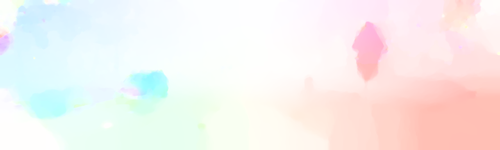} &\includegraphics[width=\linewidth]{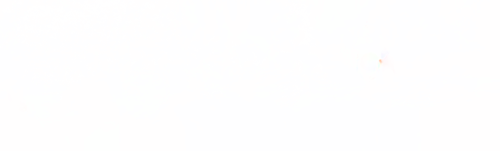} \\ 
 & & & \\[0.5em]
& Unattacked Flow & Attacked Flow & Difference
\end{tabular}
	\end{center}
	\caption{\textbf{Black-box Attacks.} Universal patch trained on FlowNet2 and PWC-Net used on all approaches pasted on a traffic sign with realistic motion as described in \refsec{sec:blackbox_attacks}.}
	\label{fig:universal_patch_attack2_RM}
\end{figure*}

We observe that the attacks on encoder-decoder networks are most effective when the patch is placed in the the center of the image. The patches influence image regions according to the receptive field of the convolutional networks. Although the patches are small ($\sim 0.5$\%) compared to the image size, they can affect up to $50$\% of the image area as can be observed in Figure \ref{fig:universal_patch_attack2_RM}.
We show more qualitative results in the supplementary material.

\subsection{Real World Attacks}
\begin{figure}
\begin{center}
	\renewcommand{\arraystretch}{0}%
	\includegraphics[width=0.95\linewidth]{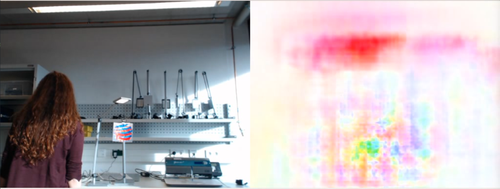} \\
	\includegraphics[width=0.95\linewidth]{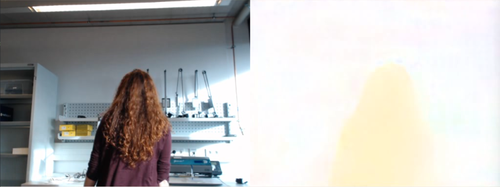}
\end{center}
\caption{\textbf{Real World Attacks on FlowNetC.} Top: The presence of a printed adversarial patch visible in the centre of the image significantly degrades the optical flow predictions. Bottom: As the patch is covered by the subject, the effect of the patch vanishes.
}
\vspace{-0.3cm}
\label{fig:real_flownetc}
\end{figure}

In order to evaluate the effectiveness of adversarial patches in the real world, we print the Black-box patch at 1200 dpi, scaled to 8 times the original size in the image.
We then place the patch in the scene, record a video sequence, and observe the effects on two networks - FlowNetC and FlowNet2.
Figure \ref{fig:real_flownetc} shows the effect of the printed patch on a deployed optical flow system running FlowNetC. While the location of the patch in the scene was constant, we experiment with and without camera motion.
We observe that the real world attacks on FlowNetC work equally well with and without camera motion, while attacks on FlowNet2 worked better with camera motion.
For the video demo of this experiment, please follow this link -- {\url{http://flowattack.is.tue.mpg.de/}}.  %

\section{Zero-Flow Test}
\label{sec:analysis}
\begin{figure*}
\begin{center}
\begin{tabularx}{\linewidth}{*{15}{@{}>{\centering\arraybackslash}X@{}}}
Input & corr & conv3\_1 & conv4 & conv5 & conv6 & flow6 & decon5 & flow5 & decon4 & flow4 & decon3 & flow3 & decon2 & predict \\
\multicolumn{15}{@{}c@{}}{\includegraphics[width=\linewidth]{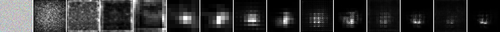}} \\
Mean & 0.0 & 0.0 & 0.0 & 0.0 & 0.0 & 0.1 & 0.1 & 0.2 & 0.2 & 0.1 & 0.2 & 0.1 & 0.1 & 0.1
\\
\multicolumn{15}{@{}c@{}}{\includegraphics[width=\linewidth]{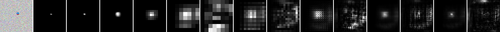}} \\
Mean & 0.1 & 6.6 & 50.3 & 98.3 & 196.5 & 464.9 &
1015.8 & 825.0 & 3035.6 & 717.8 & 2491.1 & 316.1 & 668.2 & 78.6
 \\
 \multicolumn{15}{@{}c@{}}{\textbf{FlowNetC}} \\
&&&&&&&&&&&&&&
\\
Input & corr6 & flow6 & upfeat6 & corr5 & flow5 & upfeat5 & corr4 & flow4 & upfeat4 & corr3 & flow3 & upfeat3 & corr2 & flow2
\\
\multicolumn{15}{@{}c@{}}{\includegraphics[width=\linewidth]{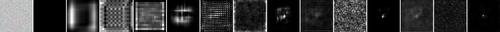}} \\
Mean & 0.0 & 12.2 & 4.4 & 0.3 & 250.9 & 157.3 & 0.0 & 31.1 & 83.0 & 0.0 & 1.5 & 14.5 & 0.0 & 0.0 \\
\multicolumn{15}{@{}c@{}}{\includegraphics[width=\linewidth]{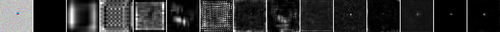}} \\
Mean & 0.00 & 12.2 & 4.4 & 0.2 & 38.2 & 134.92 & 0.04 & 2.4 & 62.6 & 0.01 & 0.1 & 13.8 & 0.00 & 0.02 \\
\multicolumn{15}{@{}c@{}}{\textbf{PWC-Net}} \\
\end{tabularx}
\end{center}
   \caption{\textbf{Visualization of Feature Maps.} Average norm of feature maps as we move deeper through the network.
   	Each image is normalized independently for contrast and scaled to the same size. Mean refers to average norm of the feature maps. }
\label{fig:feat_maps}
\end{figure*}

To better understand the behaviour of neural networks in the presence of an adversarial patch attack, we propose the \textit{Zero-Flow} test. We generate an image $I$ by independently sampling uniform random noise for each of the pixels. We then paste the universal patch $p$ on $I$, resulting in a perturbed image $\tilde{I}$.  We then replicate the image $\tilde{I}$ and input the two frames to the network. Since the image is replicated for both frames, the optical flow is zero over all the pixels.

Assuming that the attack was ineffective, $F(\tilde{I}, \tilde{I})=0$ must hold. Furthermore, if perturbation is not applied to the network inputs,  $F(I, I)=0$ must hold. For feature maps of the network, $ F_k(\tilde{I}, \tilde{I})= F_k(I,I)$ must hold. Here, $F_k$ is the output from the $k$-th layer of the neural network.
In Figure \ref{fig:feat_maps}, we show feature map visualizations along with their average norm from two networks, FlowNetC and PWC-Net. For each network we show the feature maps with and without the patch.
Each feature map is normalized independently for contrast and scaled to the same size.

Figure \ref{fig:feat_maps} illustrates that the encoder of FlowNetC (\texttt{conv<3\_1,4,5,6>}) spatially propagates the patch information, as expected, and the feature activation encompasses the entire image towards the end of the encoder \texttt{(conv6)}. While deconvolutions \texttt{(deconv<4,5,6>)} reduce the spatial propagation of activations, they also lead to a strong amplification. Finally, the flow prediction layers \texttt{(flow<5,4,3,2>)} combine both problems by fusing the encoder and decoder responses. This leads to large and widespread degradation of optical flow predictions for the Zero-Flow test.
Without the adversarial patch, we observe well distributed activations in the first convolutions as well as low activations for all layers in general.

PWC-Net shows low activations in all correlation layers \texttt{(corr<6,5,4,3,2>)} irrespective of the adversarial patch.
Interestingly, the flow decoder of lower levels \texttt{(flow<6,5>)} predict large flow in both cases -- with and without patch.
However, the flow prediction at finer levels of the pyramid \texttt{(flow<3,2>)} seems to compensate for these errors.  Similar to FlowNetC, the deconvolution layers \texttt{(upfeat<5,4,3>)} lead to an amplification of activations, with and without the patch. In addition, they seem to cause checkerboard artifacts that have also been observed in other contexts \cite{Odena2016DISTILL,Wojna2017BMVC}.

The Zero-Flow test reveals several  problems with the optical flow networks. First, the networks (FlowNetC, PWC-Net) are not spatially invariant. %
In the case of zero motion, the activation maps of the network vary spatially across the whole image even without the adversarial patch.
Second, the pyramid networks predict large motion in coarser levels \texttt{flow<5,4>} irrespective of the presence or absence of the patch.
Finally, the deconvolution layers lead to an amplification of the responses and checkerboard artifacts.
The attacks exploit these weaknesses and amplify them, degrading predictions.
In the Supplementary Material, we show additional visualizations of feature maps (from FlowNet2 and Back2Future). \blfootnote{\footnotesize
{\bf Acknowledgement.} We thank A. Punnakkal, C. Huang for paper revisions; and P. Ghosh for fruitful discussions.

{\bf \quad Disclosure.} MJB has received research
gift funds from Intel, Nvidia, Adobe, Facebook, and Amazon.
While MJB is a part-time employee of Amazon, his
research was performed solely at MPI. He is also an investor
in Meshcapde GmbH.}

\section{Discussion}
We have shown that patch attacks generalize to state-of-the-art optical flow networks and can considerably impact the performance of optical flow
systems that use deep neural networks. The patch attacks
are invariant to translation and small changes in scale and orientation of the
patch.
Learning patches for larger changes in scale and orientation is computationally costly.
Nevertheless, it is possible to replicate the attacks with real printed patterns.

Small adversarial patches cause large errors even in remote regions with encoder-decoder networks (FlowNet, FlowNet2) while spatial pyramid networks (SpyNet, PWC-Net, Back2Future) are not strongly affected.
Classical approaches are also robust to these attacks. This indicates that incorporating classical ideas such as image pyramids and correlations into architectures
makes models more robust.

In our analysis, we observed the influence of such patches on feature maps in encoder-decoder as well as spatial pyramid networks.
We also note that this results from inherent spatially variant properties of the optical flow networks. This indicates that convolutions alone are not enough to enforce spatial invariance in these networks.  %

Finally, independent of the attacks, the Zero-Flow test provides a novel approach to identify problems in deep flow networks. It reveals the internal network behaviour which may be useful to improve the robustness of flow networks.

{\small
\bibliographystyle{ieee_fullname}
\bibliography{bibliography_long,egbib}
}
\begin{appendices}
\section{Appendix}
This \textbf{supplementary document} provides additional results on White-box and Black-box attacks as well as an analysis of FlowNet2 \cite{ilg2017flownet} and Back2Future
\cite{Janai2018ECCV} under the Zero-Flow test. In the \textbf{video} \footnote{\url{http://flowattack.is.tue.mpg.de/}}, we show real world attacks using a printed patch placed in the environment.

\subsection{White-box Attacks}
\begin{figure*}[t]
\begin{center}

\begin{tabular}{@{}p{0.100\linewidth}@{}C{0.225\linewidth}@{}C{0.225\linewidth}@{}C{0.225\linewidth}@{}C{0.225\linewidth}@{}}
\rotatebox[origin=c]{30}{FlowNetC} & 
 \includegraphics[width=\linewidth]{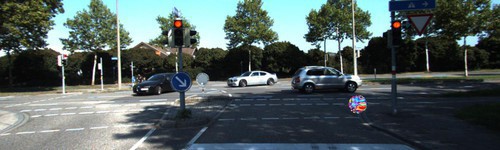} & 
 \multicolumn{3}{@{}m{\linewidth}@{}}{ \includegraphics[width=0.675\linewidth]{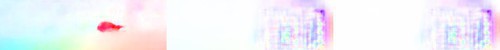}}\\ 
\rotatebox[origin=c]{30}{FlowNet2} & 
 \includegraphics[width=\linewidth]{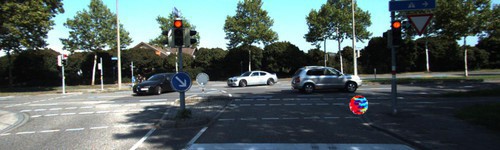} & 
 \multicolumn{3}{@{}m{\linewidth}@{}}{ \includegraphics[width=0.675\linewidth]{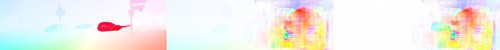}}\\ 
\rotatebox[origin=c]{30}{SpyNet} & 
 \includegraphics[width=\linewidth]{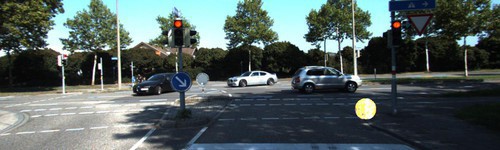} & 
 \multicolumn{3}{@{}m{\linewidth}@{}}{ \includegraphics[width=0.675\linewidth]{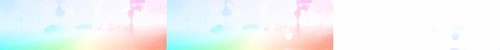}}\\ 
\rotatebox[origin=c]{30}{PWC-Net} & 
 \includegraphics[width=\linewidth]{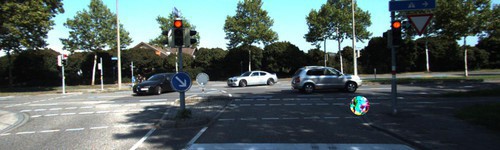} & 
 \multicolumn{3}{@{}m{\linewidth}@{}}{ \includegraphics[width=0.675\linewidth]{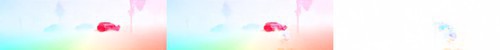}}\\ 
\rotatebox[origin=c]{30}{Back2Future} & 
 \includegraphics[width=\linewidth]{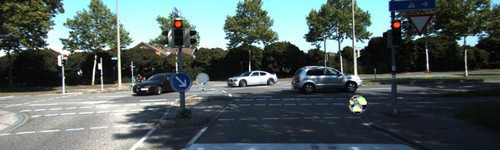} & 
 \multicolumn{3}{@{}m{\linewidth}@{}}{ \includegraphics[width=0.675\linewidth]{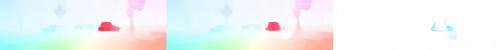}}\\ 
&&&&\\
& Attacked Reference & Unattacked Flow & Attacked Flow & Difference
\end{tabular}
\end{center}
\caption{\textbf{White-box Attacks} on all networks using 51x51 patches.}
\label{fig:51_patch_attack}
\end{figure*}

\begin{figure*}[t]
\begin{center}

\begin{tabular}{@{}p{0.100\linewidth}@{}C{0.225\linewidth}@{}C{0.225\linewidth}@{}C{0.225\linewidth}@{}C{0.225\linewidth}@{}}
\rotatebox[origin=c]{30}{FlowNetC} & 
 \includegraphics[width=\linewidth]{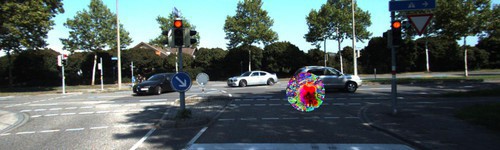} & 
 \multicolumn{3}{@{}m{\linewidth}@{}}{ \includegraphics[width=0.675\linewidth]{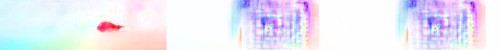}}\\ 
\rotatebox[origin=c]{30}{FlowNet2} & 
 \includegraphics[width=\linewidth]{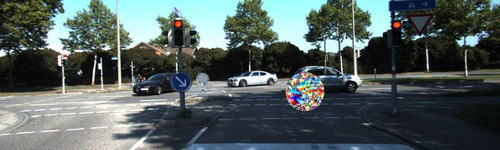} & 
 \multicolumn{3}{@{}m{\linewidth}@{}}{ \includegraphics[width=0.675\linewidth]{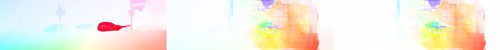}}\\ 
\rotatebox[origin=c]{30}{SpyNet} & 
 \includegraphics[width=\linewidth]{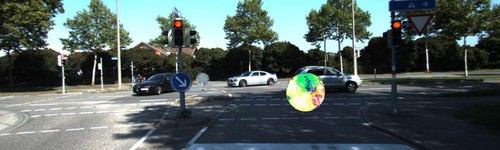} & 
 \multicolumn{3}{@{}m{\linewidth}@{}}{ \includegraphics[width=0.675\linewidth]{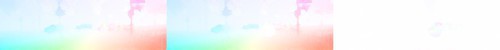}}\\ 
\rotatebox[origin=c]{30}{PWC-Net} & 
 \includegraphics[width=\linewidth]{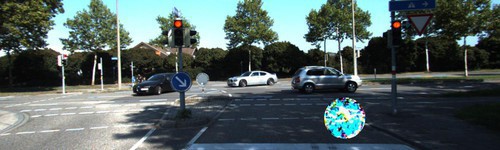} & 
 \multicolumn{3}{@{}m{\linewidth}@{}}{ \includegraphics[width=0.675\linewidth]{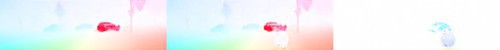}}\\ 
\rotatebox[origin=c]{30}{Back2Future} & 
 \includegraphics[width=\linewidth]{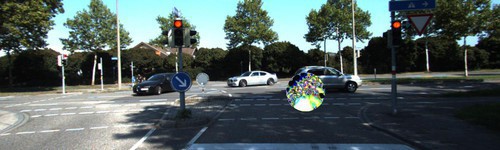} & 
 \multicolumn{3}{@{}m{\linewidth}@{}}{ \includegraphics[width=0.675\linewidth]{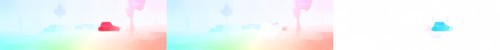}}\\ 
&&&&\\
& Attacked Reference & Unattacked Flow & Attacked Flow & Difference
\end{tabular}
\end{center}
\caption{\textbf{White-box Attacks} on all networks using 102x102 patches.}
\label{fig:102_patch_attack}
\end{figure*}

Additional qualitative results for White-box attacks using patches of size
$51 \times 51$ and $102 \times 102$ are shown in Figure
\ref{fig:51_patch_attack} and Figure \ref{fig:102_patch_attack},
respectively. We observe that the effect of the patch is more prominent with larger patch sizes. In agreement with the main paper, we note that spatial pyramid architectures are more robust, as compared to encoder-decoder architectures.

\begin{figure*}
	\begin{center}
		\includegraphics[width=\linewidth]{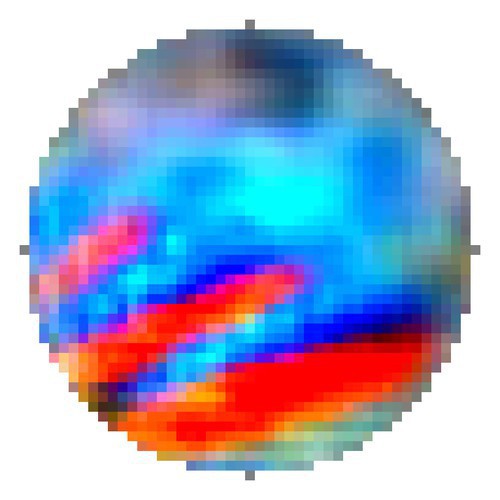}
	\end{center}
	\caption{\textbf{``Universal'' Patch} obtained by optimizing over FlowNet2 and PWCNet. Patch is enlarged for visualization.}
	\label{fig:unipatch}
\end{figure*}

\subsection{Black-box Attacks}
The universal patch is shown in Figure \ref{fig:unipatch}. Table \ref{tab:epeuni_static} shows the performance of optical flow methods when the adversarial patch has zero motion w.r.t.~the camera. In comparison to the moving Black-box attacks considered in the main paper, we observe similar effects on all networks and baselines with the adversarial patch. While encoder-decoder networks are strongly affected by the attacks, spatial pyramid networks and classical methods are more robust.

\begin{figure*}[t]
\renewcommand{\arraystretch}{0}%
\begin{center}

\begin{tabular}{@{}p{0.100\linewidth}@{}C{0.300\linewidth}@{}C{0.300\linewidth}@{}C{0.300\linewidth}@{}}
\rotatebox[origin=c]{30}{Inputs / GT} & \includegraphics[width=\linewidth]{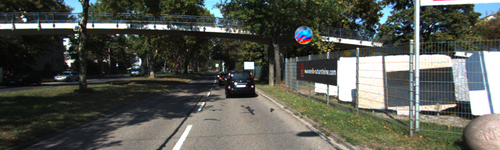} &\includegraphics[width=\linewidth]{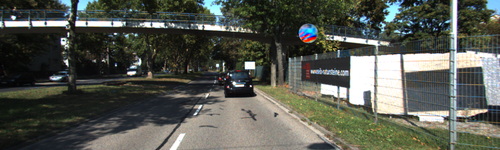} &\includegraphics[width=\linewidth]{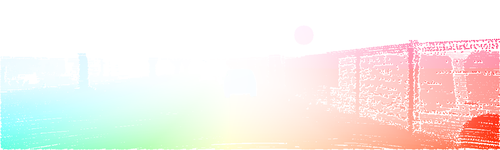} \\   
& & & \\[0.5em] \\ \rotatebox[origin=c]{30}{FlowNetC} & \multicolumn{3}{@{}m{\linewidth}@{}}{ \includegraphics[width=0.900\linewidth]{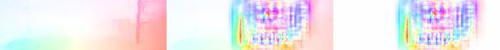}}\\ 
\rotatebox[origin=c]{30}{FlowNet2} & \multicolumn{3}{@{}m{\linewidth}@{}}{ \includegraphics[width=0.900\linewidth]{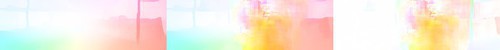}}\\ 
\rotatebox[origin=c]{30}{SpyNet} & \multicolumn{3}{@{}m{\linewidth}@{}}{ \includegraphics[width=0.900\linewidth]{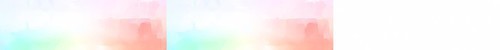}}\\ 
\rotatebox[origin=c]{30}{PWC-Net} & \multicolumn{3}{@{}m{\linewidth}@{}}{ \includegraphics[width=0.900\linewidth]{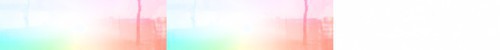}}\\ 
\rotatebox[origin=c]{30}{Back2Future} & \multicolumn{3}{@{}m{\linewidth}@{}}{ \includegraphics[width=0.900\linewidth]{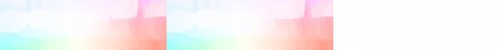}}\\ 
\rotatebox[origin=c]{30}{Epic Flow} & \includegraphics[width=\linewidth]{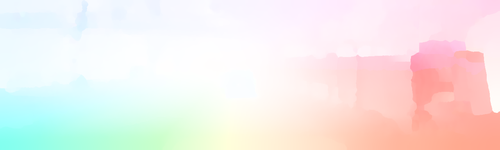} &\includegraphics[width=\linewidth]{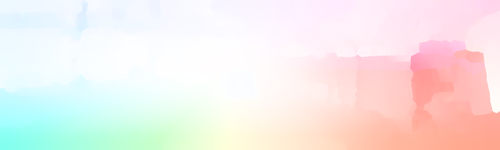} &\includegraphics[width=\linewidth]{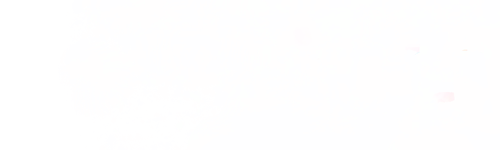} \\ 
\rotatebox[origin=c]{30}{LDOF} & \includegraphics[width=\linewidth]{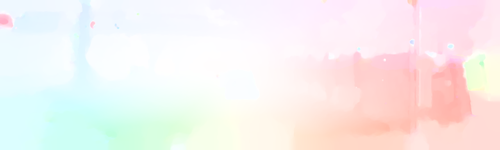} &\includegraphics[width=\linewidth]{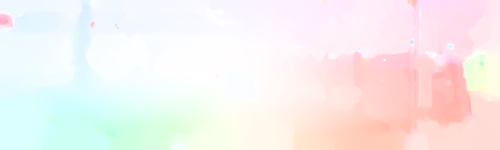} &\includegraphics[width=\linewidth]{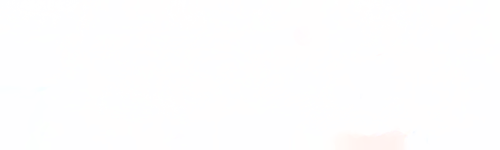} \\    
& & & \\[0.5em] 
& Unattacked Flow & Attacked Flow & Difference
\end{tabular}
\end{center}
\caption{\textbf{Black-box Attacks.} Universal patch trained on FlowNet2 and PWC-Net used on all approaches. For this evaluation, we move the patch according to the static scene.}
\label{fig:universal_patch_attack1}
\end{figure*}

\begin{figure*}[t]
\renewcommand{\arraystretch}{0}%
\begin{center}

\begin{tabular}{@{}p{0.100\linewidth}@{}C{0.300\linewidth}@{}C{0.300\linewidth}@{}C{0.300\linewidth}@{}}
\rotatebox[origin=c]{30}{Inputs / GT} & \includegraphics[width=\linewidth]{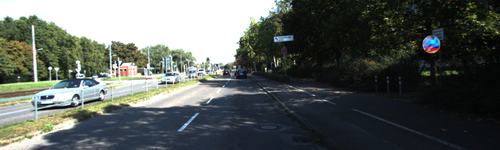} &\includegraphics[width=\linewidth]{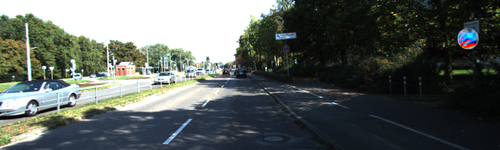} &\includegraphics[width=\linewidth]{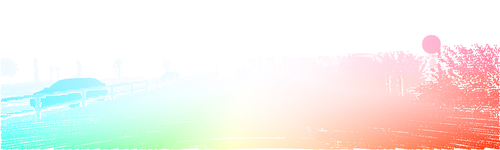} \\    
& & & \\[0.5em]  \\ \rotatebox[origin=c]{30}{FlowNetC} & \multicolumn{3}{@{}m{\linewidth}@{}}{ \includegraphics[width=0.900\linewidth]{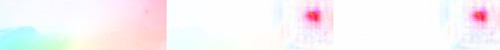}}\\ 
\rotatebox[origin=c]{30}{FlowNet2} & \multicolumn{3}{@{}m{\linewidth}@{}}{ \includegraphics[width=0.900\linewidth]{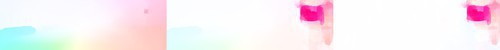}}\\ 
\rotatebox[origin=c]{30}{SpyNet} & \multicolumn{3}{@{}m{\linewidth}@{}}{ \includegraphics[width=0.900\linewidth]{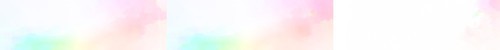}}\\ 
\rotatebox[origin=c]{30}{PWC-Net} & \multicolumn{3}{@{}m{\linewidth}@{}}{ \includegraphics[width=0.900\linewidth]{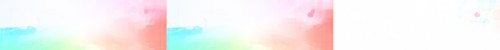}}\\ 
\rotatebox[origin=c]{30}{Back2Future} & \multicolumn{3}{@{}m{\linewidth}@{}}{ \includegraphics[width=0.900\linewidth]{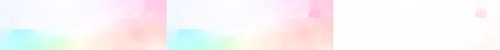}}\\ 
\rotatebox[origin=c]{30}{Epic Flow} & \includegraphics[width=\linewidth]{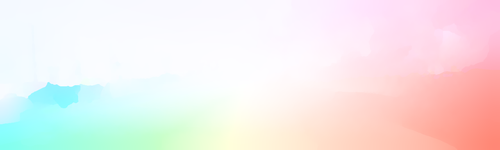} &\includegraphics[width=\linewidth]{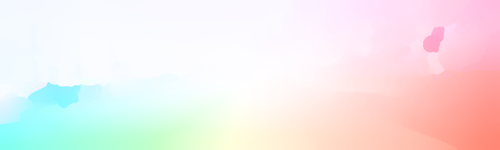} &\includegraphics[width=\linewidth]{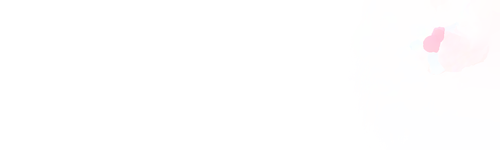} \\ 
\rotatebox[origin=c]{30}{LDOF} & \includegraphics[width=\linewidth]{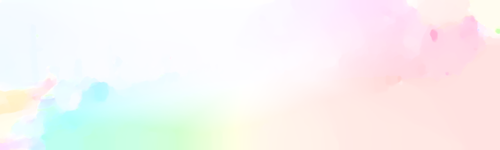} &\includegraphics[width=\linewidth]{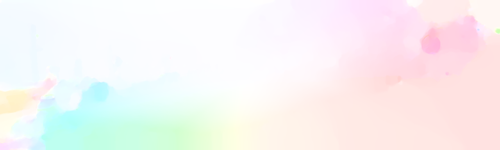} &\includegraphics[width=\linewidth]{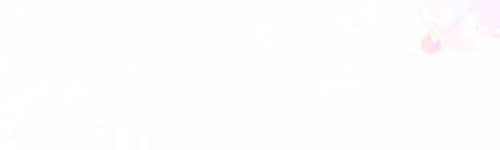} \\ 
& & & \\[0.5em] 
& Unattacked Flow & Attacked Flow & Difference
\end{tabular}
\end{center}
\caption{\textbf{Black-box Attacks.} Universal patch trained on FlowNet2 and PWC-Net used on all approaches. For this evaluation, we move the patch according to the static scene.}
\label{fig:universal_patch_attack3}
\end{figure*}

\begin{figure*}[t]
\renewcommand{\arraystretch}{0}%
\begin{center}

\begin{tabular}{@{}p{0.100\linewidth}@{}C{0.300\linewidth}@{}C{0.300\linewidth}@{}C{0.300\linewidth}@{}}
\rotatebox[origin=c]{30}{Inputs / GT} & \includegraphics[width=\linewidth]{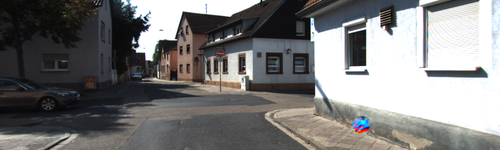} &\includegraphics[width=\linewidth]{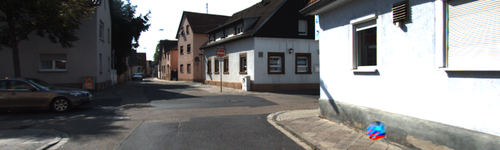} &\includegraphics[width=\linewidth]{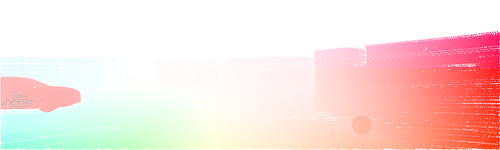} \\    
& & & \\[0.5em]  \\ \rotatebox[origin=c]{30}{FlowNetC} & \multicolumn{3}{@{}m{\linewidth}@{}}{ \includegraphics[width=0.900\linewidth]{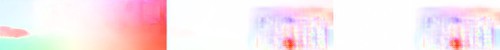}}\\ 
\rotatebox[origin=c]{30}{FlowNet2} & \multicolumn{3}{@{}m{\linewidth}@{}}{ \includegraphics[width=0.900\linewidth]{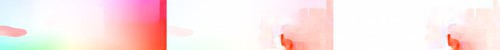}}\\ 
\rotatebox[origin=c]{30}{SpyNet} & \multicolumn{3}{@{}m{\linewidth}@{}}{ \includegraphics[width=0.900\linewidth]{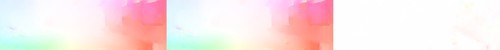}}\\ 
\rotatebox[origin=c]{30}{PWC-Net} & \multicolumn{3}{@{}m{\linewidth}@{}}{ \includegraphics[width=0.900\linewidth]{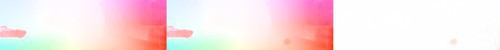}}\\ 
\rotatebox[origin=c]{30}{Back2Future} & \multicolumn{3}{@{}m{\linewidth}@{}}{ \includegraphics[width=0.900\linewidth]{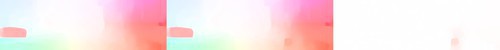}}\\ 
\rotatebox[origin=c]{30}{Epic Flow} & \includegraphics[width=\linewidth]{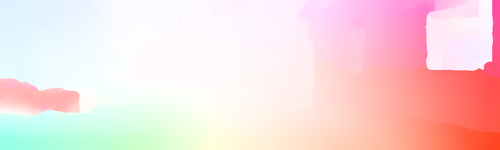} &\includegraphics[width=\linewidth]{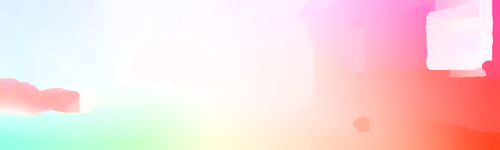} &\includegraphics[width=\linewidth]{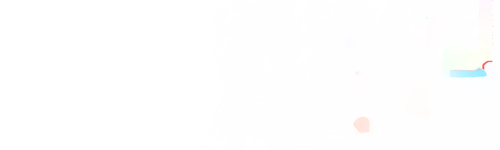} \\ 
\rotatebox[origin=c]{30}{LDOF} & \includegraphics[width=\linewidth]{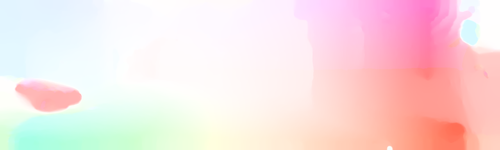} &\includegraphics[width=\linewidth]{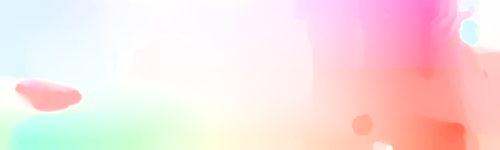} &\includegraphics[width=\linewidth]{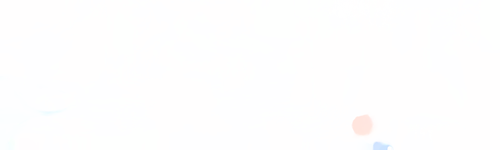} \\    
& & & \\[0.5em] 
& Unattacked Flow & Attacked Flow & Difference
\end{tabular}
\end{center}
\caption{\textbf{Black-box Attacks.} Universal patch trained on FlowNet2 and PWC-Net used on all approaches. For this evaluation, we move the patch according to the static scene.}
\label{fig:universal_patch_attack4}
\end{figure*}

\begin{figure*}[t]
\renewcommand{\arraystretch}{0}%
\begin{center}

\begin{tabular}{@{}p{0.100\linewidth}@{}C{0.300\linewidth}@{}C{0.300\linewidth}@{}C{0.300\linewidth}@{}}
\rotatebox[origin=c]{30}{Inputs / GT} & \includegraphics[width=\linewidth]{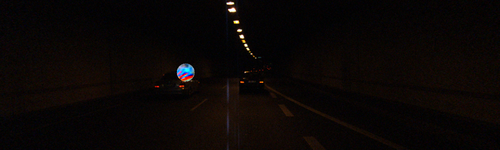} &\includegraphics[width=\linewidth]{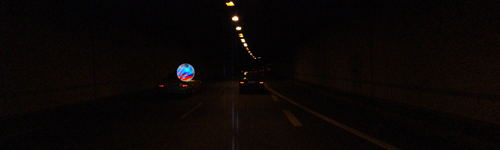} &\includegraphics[width=\linewidth]{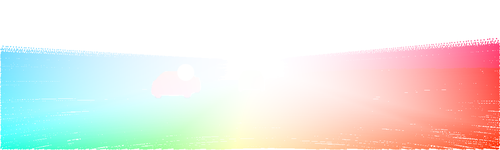} \\    
& & & \\[0.5em]  \\ \rotatebox[origin=c]{30}{FlowNetC} & \multicolumn{3}{@{}m{\linewidth}@{}}{ \includegraphics[width=0.900\linewidth]{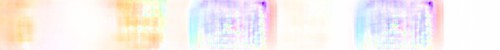}}\\ 
\rotatebox[origin=c]{30}{FlowNet2} & \multicolumn{3}{@{}m{\linewidth}@{}}{ \includegraphics[width=0.900\linewidth]{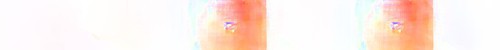}}\\ 
\rotatebox[origin=c]{30}{SpyNet} & \multicolumn{3}{@{}m{\linewidth}@{}}{ \includegraphics[width=0.900\linewidth]{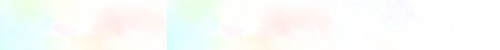}}\\ 
\rotatebox[origin=c]{30}{PWC-Net} & \multicolumn{3}{@{}m{\linewidth}@{}}{ \includegraphics[width=0.900\linewidth]{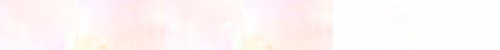}}\\ 
\rotatebox[origin=c]{30}{Back2Future} & \multicolumn{3}{@{}m{\linewidth}@{}}{ \includegraphics[width=0.900\linewidth]{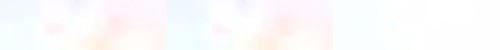}}\\ 
\rotatebox[origin=c]{30}{Epic Flow} & \includegraphics[width=\linewidth]{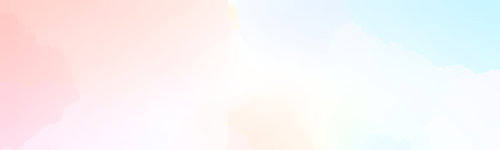} &\includegraphics[width=\linewidth]{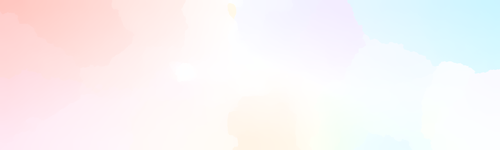} &\includegraphics[width=\linewidth]{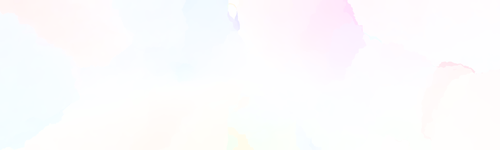} \\ 
\rotatebox[origin=c]{30}{LDOF} & \includegraphics[width=\linewidth]{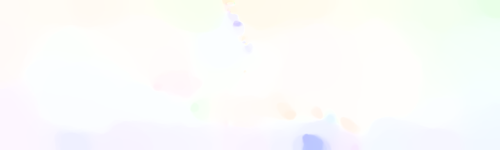} &\includegraphics[width=\linewidth]{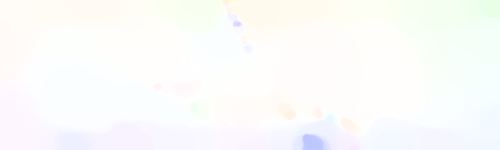} &\includegraphics[width=\linewidth]{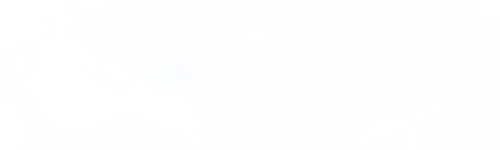} \\    
& & & \\[0.5em] 
& Unattacked Flow & Attacked Flow & Difference
\end{tabular}
\caption{\textbf{Black-box Attacks.} Universal patch trained on FlowNet2 and PWC-Net used on all approaches. For this evaluation, we move the patch according to the static scene.}
\label{fig:universal_patch_attack5}
\end{center}
\end{figure*}

\begin{figure*}[t]
\renewcommand{\arraystretch}{0}%
\begin{center}

\begin{tabular}{@{}p{0.100\linewidth}@{}C{0.300\linewidth}@{}C{0.300\linewidth}@{}C{0.300\linewidth}@{}}
\rotatebox[origin=c]{30}{Inputs / GT} & \includegraphics[width=\linewidth]{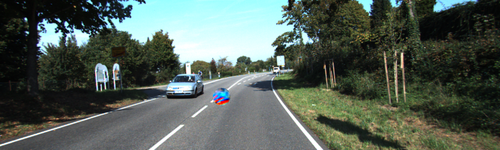} &\includegraphics[width=\linewidth]{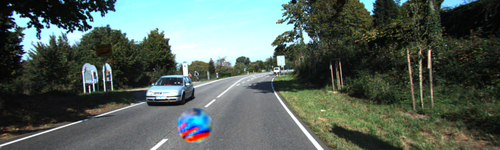} &\includegraphics[width=\linewidth]{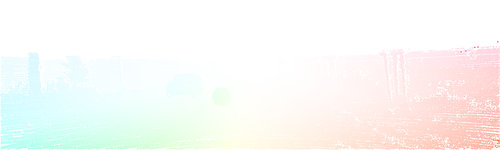} \\    
& & & \\[0.5em]  \\ \rotatebox[origin=c]{30}{FlowNetC} & \multicolumn{3}{@{}m{\linewidth}@{}}{ \includegraphics[width=0.900\linewidth]{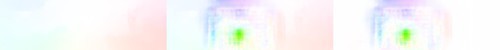}}\\ 
\rotatebox[origin=c]{30}{FlowNet2} & \multicolumn{3}{@{}m{\linewidth}@{}}{ \includegraphics[width=0.900\linewidth]{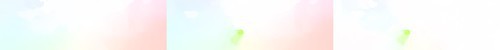}}\\ 
\rotatebox[origin=c]{30}{SpyNet} & \multicolumn{3}{@{}m{\linewidth}@{}}{ \includegraphics[width=0.900\linewidth]{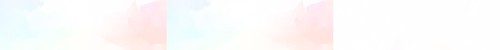}}\\ 
\rotatebox[origin=c]{30}{PWC-Net} & \multicolumn{3}{@{}m{\linewidth}@{}}{ \includegraphics[width=0.900\linewidth]{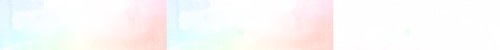}}\\ 
\rotatebox[origin=c]{30}{Back2Future} & \multicolumn{3}{@{}m{\linewidth}@{}}{ \includegraphics[width=0.900\linewidth]{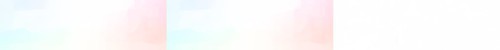}}\\ 
\rotatebox[origin=c]{30}{Epic Flow} & \includegraphics[width=\linewidth]{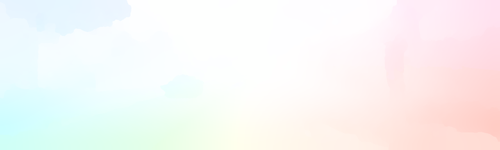} &\includegraphics[width=\linewidth]{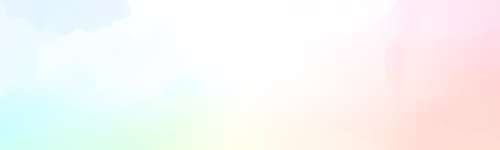} &\includegraphics[width=\linewidth]{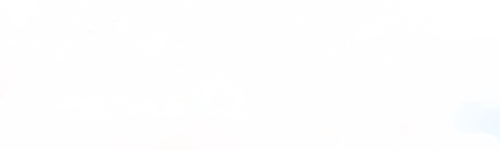} \\ 
\rotatebox[origin=c]{30}{LDOF} & \includegraphics[width=\linewidth]{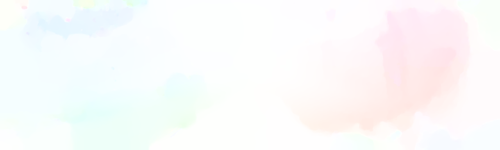} &\includegraphics[width=\linewidth]{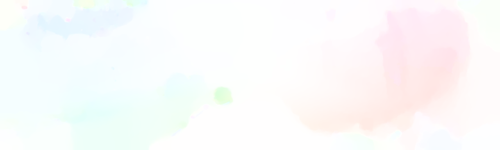} &\includegraphics[width=\linewidth]{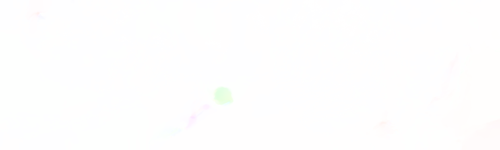} \\    
& & & \\[0.5em] 
& Unattacked Flow & Attacked Flow & Difference
\end{tabular}
\caption{\textbf{Black-box Attacks.} Universal patch trained on FlowNet2 and PWC-Net used on all approaches. For this evaluation, we move the patch according to the static scene.}
\label{fig:universal_patch_attack6}
\end{center}
\end{figure*}

\begin{figure*}[t]
\renewcommand{\arraystretch}{0}%
\begin{center}

\begin{tabular}{@{}p{0.100\linewidth}@{}C{0.300\linewidth}@{}C{0.300\linewidth}@{}C{0.300\linewidth}@{}}
\rotatebox[origin=c]{30}{Inputs / GT} & \includegraphics[width=\linewidth]{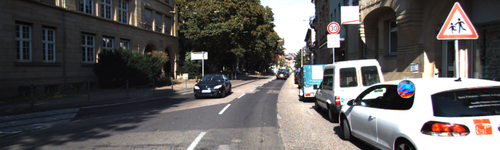} &\includegraphics[width=\linewidth]{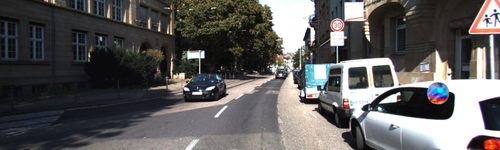} &\includegraphics[width=\linewidth]{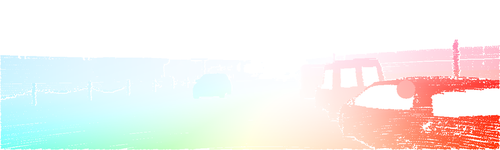} \\    
& & & \\[0.5em]  \\ \rotatebox[origin=c]{30}{FlowNetC} & \multicolumn{3}{@{}m{\linewidth}@{}}{ \includegraphics[width=0.900\linewidth]{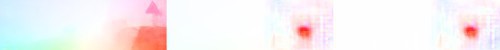}}\\ 
\rotatebox[origin=c]{30}{FlowNet2} & \multicolumn{3}{@{}m{\linewidth}@{}}{ \includegraphics[width=0.900\linewidth]{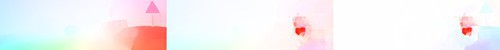}}\\ 
\rotatebox[origin=c]{30}{SpyNet} & \multicolumn{3}{@{}m{\linewidth}@{}}{ \includegraphics[width=0.900\linewidth]{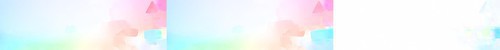}}\\ 
\rotatebox[origin=c]{30}{PWC-Net} & \multicolumn{3}{@{}m{\linewidth}@{}}{ \includegraphics[width=0.900\linewidth]{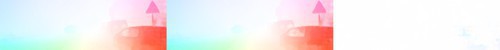}}\\ 
\rotatebox[origin=c]{30}{Back2Future} & \multicolumn{3}{@{}m{\linewidth}@{}}{ \includegraphics[width=0.900\linewidth]{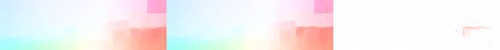}}\\ 
\rotatebox[origin=c]{30}{Epic Flow} & \includegraphics[width=\linewidth]{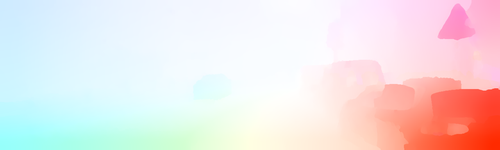} &\includegraphics[width=\linewidth]{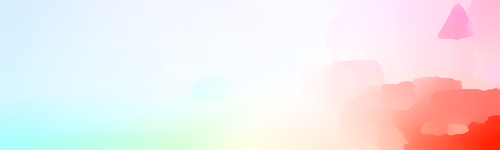} &\includegraphics[width=\linewidth]{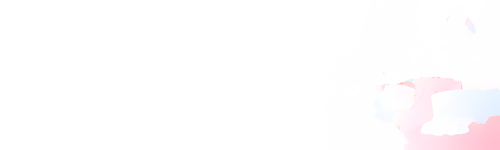} \\ 
\rotatebox[origin=c]{30}{LDOF} & \includegraphics[width=\linewidth]{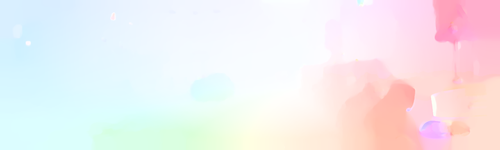} &\includegraphics[width=\linewidth]{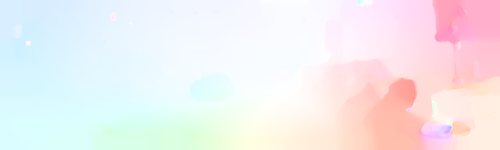} &\includegraphics[width=\linewidth]{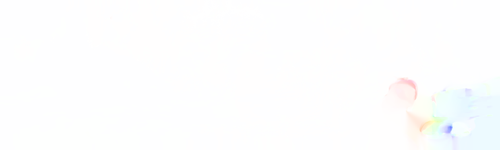} \\    
& & & \\[0.5em] 
& Unattacked Flow & Attacked Flow & Difference
\end{tabular}
\caption{\textbf{Black-box Attacks.} Universal patch trained on FlowNet2 and PWC-Net used on all approaches. For this evaluation, we move the patch according to the static scene.}
\label{fig:universal_patch_attack7}
\end{center}
\end{figure*}

\begin{figure*}[t]
\renewcommand{\arraystretch}{0}%
\begin{center}

\begin{tabular}{@{}p{0.100\linewidth}@{}C{0.300\linewidth}@{}C{0.300\linewidth}@{}C{0.300\linewidth}@{}}
\rotatebox[origin=c]{30}{Inputs / GT} & \includegraphics[width=\linewidth]{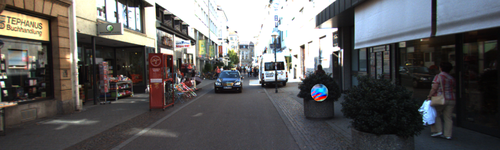} &\includegraphics[width=\linewidth]{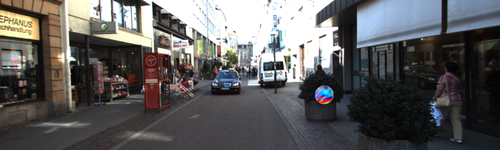} &\includegraphics[width=\linewidth]{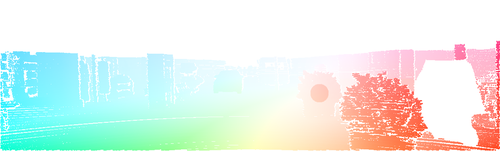} \\    
& & & \\[0.5em]  \\ \rotatebox[origin=c]{30}{FlowNetC} & \multicolumn{3}{@{}m{\linewidth}@{}}{ \includegraphics[width=0.900\linewidth]{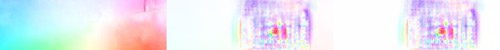}}\\ 
\rotatebox[origin=c]{30}{FlowNet2} & \multicolumn{3}{@{}m{\linewidth}@{}}{ \includegraphics[width=0.900\linewidth]{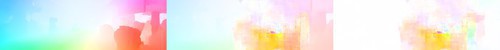}}\\ 
\rotatebox[origin=c]{30}{SpyNet} & \multicolumn{3}{@{}m{\linewidth}@{}}{ \includegraphics[width=0.900\linewidth]{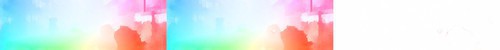}}\\ 
\rotatebox[origin=c]{30}{PWC-Net} & \multicolumn{3}{@{}m{\linewidth}@{}}{ \includegraphics[width=0.900\linewidth]{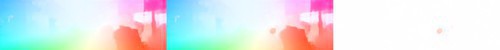}}\\ 
\rotatebox[origin=c]{30}{Back2Future} & \multicolumn{3}{@{}m{\linewidth}@{}}{ \includegraphics[width=0.900\linewidth]{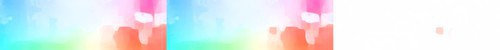}}\\ 
\rotatebox[origin=c]{30}{Epic Flow} & \includegraphics[width=\linewidth]{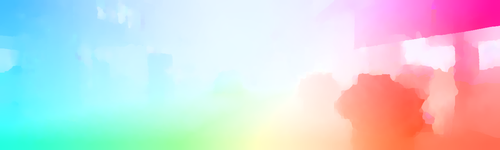} &\includegraphics[width=\linewidth]{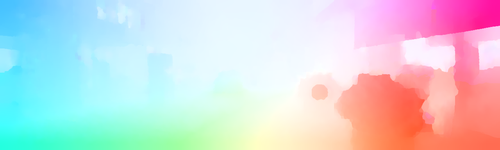} &\includegraphics[width=\linewidth]{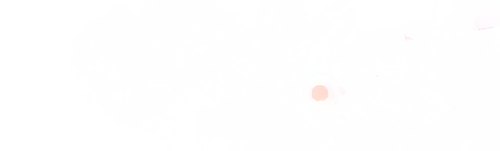} \\ 
\rotatebox[origin=c]{30}{LDOF} & \includegraphics[width=\linewidth]{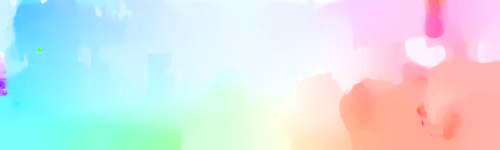} &\includegraphics[width=\linewidth]{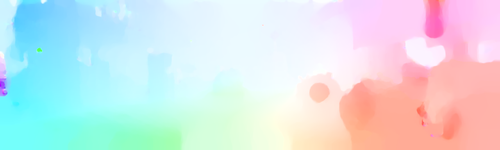} &\includegraphics[width=\linewidth]{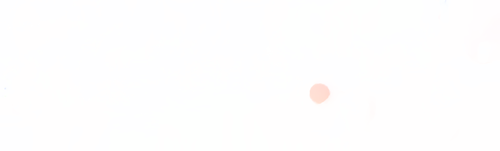} \\    
& & & \\[0.5em] 
& Unattacked Flow & Attacked Flow & Difference
\end{tabular}
\caption{\textbf{Black-box Attacks.} Universal patch trained on FlowNet2 and PWC-Net used on all approaches. For this evaluation, we move the patch according to the static scene.}
\label{fig:universal_patch_attack8}
\end{center}
\end{figure*}

In Figures \ref{fig:universal_patch_attack1} -
\ref{fig:universal_patch_attack8} we show some additional qualitative
results for the Black-box attack with patches moving according to the
scene as described in the main paper. These examples demonstrate the feasibility of such attackes in the real world.
In Figure \ref{fig:universal_patch_attack3}, for instance, the patch is attached to and moves with a traffic sign, while Figures \ref{fig:universal_patch_attack4}, \ref{fig:universal_patch_attack7} illustrate cases when a patch is printed on a wall and a car.

\begin{table}[t]
\begin{center}
\begin{tabular}{l|ccc}
    & Unattacked & \multicolumn{2}{c}{Attacked}    \\
        &  EPE   &EPE  &  Rel\\
\hline
    FlowNet2 \cite{ilg2017flownet} &       11.90 &     30.99 &     +160 \% \\
      PWCNet \cite{Sun2018CVPR} &       11.03 &     11.16 &       +1
                                                            \%  \\ \hline
    FlowNetC \cite{dosovitskiy2015flownet} &       14.56 &     77.78 &     +434 \% \\
      SpyNet \cite{spynet2017} &       20.26 &     20.65 &       +2 \% \\
 Back2Future \cite{Janai2018ECCV} &       17.49 &     17.76 &       +2 \% \\

   Epic Flow \cite{Revaud2015CVPR} &        4.52 &      4.57 &       +1 \% \\
        LDOF \cite{Brox2011PAMI} &        9.20 &      9.30 &       +1 \% \\
\end{tabular}

\caption{\textbf{Black-box Attacks.} Attacks on different optical flow methods using a universal patch that is static w.r.t.~the camera. Methods below the line were not used for training the patch.}
\label{tab:epeuni_static}
\end{center}
\end{table}

\paragraph{Evaluation without considering the Patch Region.}
We also evaulated the effect of the patch without considering the patch region.
In case of Black-box attacks (\tabref{tab:epeuni2}) the flow outside of the patch region has a similar level of degradation as our results considering the patch region.
The unattacked results only show minimal changes below the second decimal place because of the small patch size ($\approx 1\%$).

\begin{table}[t]
	\begin{center}
		\resizebox{\linewidth}{!}{\begin{tabular}{l|cc|cc}
&  \multicolumn{2}{c|}{Unattacked} &  \multicolumn{2}{c}{Attacked}  \\
                                        &  W Patch &  W/O Patch &   W Patch &  W/O Patch \\
\hline
FlowNetC  &       14.56 & 14.56 &       86.12 &    80.69 \\
PWCNet &       11.03 &   11.03 &      11.01 &     11.08 \\
	FlowNet2 &       11.90 &     11.90 &    36.13 &     34.18 \\
     SpyNet  &      20.26 &     20.26 &      20.39 &    20.50 \\
Back2Future  &     17.49 &     17.49 &    17.44 &      17.59 \\
\end{tabular}
}
		\kern1.0em
		\caption{\textbf{Black-box Attacks.} Comparison of the evaluation results with and without considering the attacking patch region.}
		\label{tab:epeuni2}
	\end{center}
\kern-1.5em
\end{table}


\subsection{Zero-Flow Test}
\begin{figure*}
	\begin{center}
	\begin{tabularx}{\linewidth}{*{15}{@{}>{\centering\arraybackslash}X@{}}}
	Input & corr6 & flow6 & upfeat6 & corr5 & flow5 & upfeat5 & corr4 & flow4 & upfeat4 & corr3 & flow3 & upfeat3 & corr2 & flow2
	\end{tabularx}
	\includegraphics[width=\linewidth]{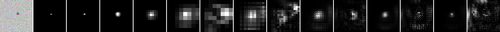}
	\begin{tabularx}{\linewidth}{*{15}{@{}>{\centering\arraybackslash}X@{}}}
	Mean & 0.0 & 	3.7 & 	23.6 & 	51.7 &	54.9 &	207.5 &	229.7 &	235.6 &	677.6 &
	221.1 & 	490.5 &	75.4 &	119.2 &	34.2 \\
	\end{tabularx}
	\includegraphics[width=\linewidth]{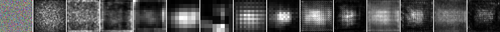}
	\begin{tabularx}{\linewidth}{*{15}{@{}>{\centering\arraybackslash}X@{}}}
	Mean & 0.0 & 0.0 & 0.0 & 0.0 & 0.0 & 0.0 & 0.0 & 0.0 & 0.0 & 0.0 & 0.1 & 0.0 & 0.1 & 0.0  \\
	\multicolumn{15}{@{}c@{}}{ \textbf{FlowNet2 (FlowNetC)}} \\
	\end{tabularx}

	\begin{tabularx}{\linewidth}{*{15}{@{}>{\centering\arraybackslash}X@{}}}
	conv1 & conv2 & conv3 & conv4 & conv5 & conv6 & flow6 &	deconv5 & flow5 & deconv4 & flow4 & deconv3 & flow3 & deconv2 & flow2
	\end{tabularx}
	\includegraphics[width=\linewidth]{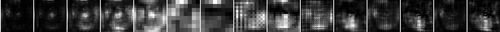}
	\begin{tabularx}{\linewidth}{*{15}{@{}>{\centering\arraybackslash}X@{}}}
	3.0 & 2.8 & 3.0 & 1.2 & 1.0 & 0.2 & 6.7 & 1.7 & 8.4 & 16.5 & 12.3 & 32.6 & 10.4 & 39.9 & 10.6  \\
	\end{tabularx}
	\includegraphics[width=\linewidth]{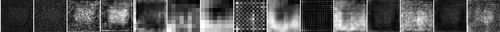}
	\begin{tabularx}{\linewidth}{*{15}{@{}>{\centering\arraybackslash}X@{}}}
	0.3 & 0.1  & 0.0  & 0.0  & 0.0  & 0.0  & 0.0  & 0.0  & 0.0  & 0.0  & 0.0  & 0.0  & 0.0  & 0.0  & 0.0 \\
	\multicolumn{15}{@{}c@{}}{ \textbf{FlowNet2 (FlowNetS)}} \\
	\end{tabularx}

	\begin{tabularx}{\linewidth}{*{15}{@{}>{\centering\arraybackslash}X@{}}}
	conv1 & conv2 & conv3 & conv4 & conv5 & conv6 & flow6 &	deconv5 & flow5 & deconv4 & flow4 & deconv3 & flow3 & deconv2 & flow2
	\end{tabularx}
	\includegraphics[width=\linewidth]{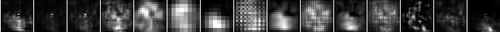}
	\begin{tabularx}{\linewidth}{*{15}{@{}>{\centering\arraybackslash}X@{}}}
	2.6 & 0.6 & 0.1 & 0.1 & 0.0 & 0.0 & 0.1 & 0.0 & 0.9 & 0.2 & 1.2 & 1.1 & 0.8 & 3.0 & 3.0 \\
	\end{tabularx}
	\includegraphics[width=\linewidth]{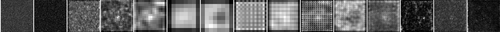}
	\begin{tabularx}{\linewidth}{*{15}{@{}>{\centering\arraybackslash}X@{}}}
	0.3 & 0.0 & 0.0 & 0.0 & 0.0 & 0.0 & 0.0 & 0.0 & 0.0 & 0.0 & 0.0 & 0.0 & 0.0 & 0.0 & 0.0 \\
	\multicolumn{15}{@{}c@{}}{ \textbf{FlowNet2 (FlowNetS)}} \\
	\end{tabularx}

	\begin{tabularx}{\linewidth}{*{16}{@{}>{\centering\arraybackslash}X@{}}}
	{\small conv0} & {\small conv1} & {\small conv2} & {\small conv3} & {\small conv4} & {\small conv5} & {\small conv6} & {\small flow6} & {\small deconv5} & {\small flow5} & {\small deconv4} & {\small flow4} & {\small deconv3} & {\small flow3} & {\small deconv2} & {\small flow2} \\
	\end{tabularx}
	\includegraphics[width=\linewidth]{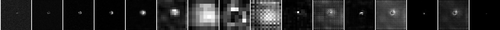} \\
	\begin{tabularx}{\linewidth}{*{16}{@{}>{\centering\arraybackslash}X@{}}}
	0.3 & 	0.1 & 	0.0 & 	0.0 & 	0.1 & 	1.3 & 	12.4 & 	0.0 & 	8.5 & 	0.0 & 	11.6 & 	0.0 & 	12.4 & 	0.0 & 	1.0 & 	0.0  \\
	\end{tabularx}
	\includegraphics[width=\linewidth]{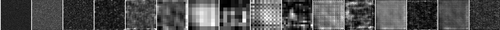} \\
	\begin{tabularx}{\linewidth}{*{16}{@{}>{\centering\arraybackslash}X@{}}}
   0.3 & 	0.1 & 	0.0 & 	0.0 & 	0.1 & 	1.2 & 	11.5 & 	0.0 & 	7.7 & 	0.0 & 	11.0 & 	0.0 & 	11.9 & 	0.0 & 	1.0 & 	0.0  \\
	 \multicolumn{16}{@{}c@{}}{ \textbf{FlowNet2 (FlowNetSD)}} \\
	\end{tabularx}

	\begin{tabularx}{0.6\linewidth}{*{8}{@{}>{\centering\arraybackslash}X@{}}}
	conv0 & conv1 & conv2 & flow2 &  deconv1 & flow1 & deconv0 & flow0 \\
	\end{tabularx}
	\includegraphics[width=0.6\linewidth]{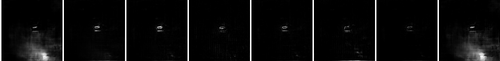} \\
	\begin{tabularx}{0.6\linewidth}{*{8}{@{}>{\centering\arraybackslash}X@{}}}
	186.9 & 8.3 & 39.8 & 99.8 & 113.2 & 276.6 & 104.1 & 1221.9 \\
	\end{tabularx}
	\includegraphics[width=0.6\linewidth]{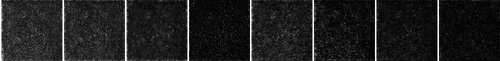} \\
	\begin{tabularx}{0.6\linewidth}{*{8}{@{}>{\centering\arraybackslash}X@{}}}
	0.1 & 0.0 & 0.1 & 0.4 & 0.4 & 1.0 & 0.5 & 0.0 \\
	\multicolumn{8}{@{}c@{}}{ \textbf{FlowNet2 (FlowNet Fusion)}} \\
	\end{tabularx}
	\end{center}

	\caption{\textbf{Zero-Flow Test.} Feature maps of Flownet2 under the Zero-Flow test. Top to bottom, we show rows corresponding to FlowNetC, FlowNetS, FlowNetS, FlowNetSD and FlowNet Fusion that constitute FlowNet2.}
	\label{fig:feat_flownet2}
\end{figure*}

\begin{figure*}
	\begin{tabularx}{\linewidth}{*{15}{@{}>{\centering\arraybackslash}X@{}}}
	Input &	corr6 & flow6 & corr5 & upfeat5 & flow5 & corr4 & upfeat4 & flow4 &
									corr3 & upfeat3 & flow3 & corr2 & upfeat2 & flow2 \\
	\multicolumn{15}{@{}c@{}}{\includegraphics[width=\linewidth]{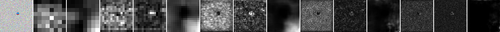}} \\
	Mean & 0.0 & 0.2 & 0.0 & 0.2 & 0.0 & 0.0 & 0.1 & 0.0 & 0.0 & 0.0 & 0.0 & 0.0 & 0.0 & 0.0 \\

	\multicolumn{15}{@{}c@{}}{\includegraphics[width=\linewidth]{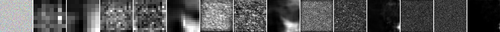}} \\
	Mean & 0.0 & 0.2 & 0.0 & 0.2 & 0.0 & 0.0 & 0.0 & 0.0 & 0.0 & 0.0 & 0.0 & 0.0 & 0.0 & 2.0 \\

	& & & & & & & \textbf{Forward} \\ \\
	Input &	corr6 & flow6 & corr5 & upfeat5 & flow5 & corr4 & upfeat4 & flow4 &
									corr3 & upfeat3 & flow3 & corr2 & upfeat2 & flow2 \\

	\multicolumn{15}{@{}c@{}}{\includegraphics[width=\linewidth]{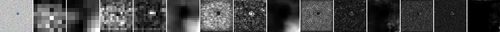}} \\
	Mean & 0.0 & 0.2 & 0.0 & 0.2 & 0.0 & 0.0 & 0.0 & 0.0 & 0.0 & 0.0 & 0.0 & 0.0 & 0.0 & 0.0 \\

	\multicolumn{15}{@{}c@{}}{\includegraphics[width=\linewidth]{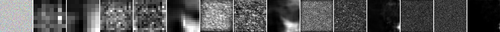}} \\
	Mean & 0.0 & 0.2 & 0.0 & 0.2 & 0.0 & 0.0 & 0.1 & 0.0 & 0.0 & 0.0 & 0.0 & 0.0 & 0.0 & 0.0 \\
	& & & & & & &	\textbf{Backward}
	\end{tabularx}

	\caption{\textbf{Zero-Flow Test.} Feature maps of Back2Future under Zero-Flow test. Top to bottom, we show rows corresponding to forward and backward parts of Back2Future in a multi-frame set up.}
	\label{fig:feat_back2future}
\end{figure*}

We show feature map visualizations for FlowNet2 and Back2Future under
the Zero-Flow test in Figures \ref{fig:feat_flownet2} and
\ref{fig:feat_back2future} respectively. We note that the feature maps
of FlowNet2 are not spatially invariant, which is consistant with other
networks examined in Section 5 of the main paper.
The stacked FlowNetS (part of FlowNet2) seems to be less vulnerable to the adversarial patch as compared to FlowNetC (part of FlowNet2). We also observe that the fusion part of FlowNet2 dramatically amplifies the degradations in optical flow predictions. The deconvolution layers show similar checkerboard artifacts as FlowNetC and PWC-Net analysed in Section 5 of the main paper.

For Back2Future, we note that, although the feature maps are not
spatially invariant, their magnitude remains small irrespective of the
presence or absence of the adversary. Interestingly, Back2Future gives reasonable flow predictions at coarser levels of the pyramid unlike PWC-Net, even though they share a common architecture.

We note that the problem of spatially variant feature maps continue across all the examined networks, along with the checkerboard artifacts.

\end{appendices}

\end{document}